\documentclass[pdflatex,sn-mathphys-num]{sn-jnl}

\usepackage{subcaption}
\usepackage{pifont}

\usepackage{graphicx}%
\usepackage{multirow}%
\usepackage{amsmath,amssymb,amsfonts}%
\usepackage{amsthm}%
\usepackage{mathrsfs}%
\usepackage[title]{appendix}%
\usepackage{xcolor}%
\usepackage{textcomp}%
\usepackage{manyfoot}%
\usepackage{booktabs}%
\usepackage{algorithm}%
\usepackage{algorithmicx}%
\usepackage{algpseudocode}%
\usepackage{listings}%

\usepackage{lscape}  
\usepackage{afterpage}  

\theoremstyle{thmstyleone}%
\theoremstyle{thmstyletwo}%

\theoremstyle{thmstylethree}%

\begin{document}

\title[LLMs to bootstrap reinforcement learning]{Can we use LLMs to bootstrap reinforcement learning? - A case study in digital health behavior change}


\author*[1]{\fnm{Nele} \sur{Albers}}
\author[2]{\fnm{Esra Cemre Su} \sur{de Groot}}
\author[3]{\fnm{Loes} \sur{Keijsers}}
\author[4]{\fnm{Manon H.} \sur{Hillegers}}
\author[1]{\fnm{Emiel} \sur{Krahmer}}

\affil*[1]{\orgdiv{Department of Communication and Cognition}, \orgname{Tilburg University}, \orgaddress{\city{Tilburg}, \country{Netherlands}}}
\affil[2]{\orgdiv{Department of Software Technology}, \orgname{Delft University of Technology}, \orgaddress{\city{Delft}, \country{Netherlands}}}
\affil[3]{\orgdiv{Department of Psychology, Education and Child Studies}, \orgname{Erasmus University Rotterdam}, \orgaddress{\city{Rotterdam}, \country{Netherlands}}}
\affil[4]{\orgdiv{Department of Child and Adolescent Psychiatry/Psychology}, \orgname{Erasmus MC Sophia Children's Hospital}, \orgaddress{\city{Rotterdam}, \country{Netherlands}}}


\abstract{Personalizing digital applications for health behavior change is a promising route to making them more engaging and effective. This especially holds for approaches that adapt to users and their specific states (e.g., motivation, knowledge, wants) over time. However, developing such approaches requires making many design choices, whose effectiveness is difficult to predict from literature and costly to evaluate in practice. In this work, we explore whether large language models (LLMs) can be used out-of-the-box to generate samples of user interactions that provide useful information for training reinforcement learning models for digital behavior change settings. Using real user data from four large behavior change studies as comparison, we show that LLM-generated samples can be useful in the absence of real data. Comparisons to the samples provided by human raters further show that LLM-generated samples reach the performance of human raters. Additional analyses of different prompting strategies including shorter and longer prompt variants, chain-of-thought prompting, and few-shot prompting show that the relative effectiveness of different strategies depends on both the study and the LLM with also relatively large differences between prompt paraphrases alone. We provide recommendations for how LLM-generated samples can be useful in practice.}


\maketitle

\section*{Introduction}

While effective for changing various health behaviors including ones related to lifestyle \cite{singh2024systematic} and mental well-being \cite{groot2023effectiveness}, digital health applications likely work best when they can be personalized and adapt to the individual wishes and needs of users whenever they arise \cite{kocaballi2019personalization,singh2024systematic}. Technically, this kind of personalized interactions can be achieved using reinforcement learning (RL) \cite{sutton2018reinforcement}, a machine learning approach where an intelligent agent learns to make decisions by taking actions, observing the impact these actions have on users, and adjusting its behavior accordingly. Crucially, this method relies on data on users, in the form of rewards and transitions between user states, to determine the impact actions have. But what if no human user data is available? In this paper we explore whether large language models (LLMs) can be used to generate useful interaction samples for digital health behavior change settings.

LLMs have recently shown promise in simulating patients in conversations with mental health trainees \cite{wang2024patient}, nursing students \cite{agatsuma2024building}, and medical students \cite{brugge2024large}. This means that even though LLMs are trained to predict the next word for a given text, they have the potential to generate cognition and behavior simulating those of humans. While some approaches explicitly integrate cognitive models with LLMs (e.g., \cite{wang2024patient}), others use relatively simple prompts (e.g., \cite{agatsuma2024building,brugge2024large}). The latter highlights the potential of LLMs to generate interaction samples more or less out of the box, without the need to carefully craft simulations based on data from related studies (e.g., \cite{li2024dyadicreinforcementlearning,trella2024oralyticsreinforcementlearningalgorithm,coughlin2024amobile}) or try to derive expert- or theory-based rules which are commonly not available at a sufficiently detailed level \cite{klasnja2015microrandomized}.

To examine whether it is beneficial to use out-of-the-box LLMs to simulate the behavior of users in digital behavior change applications, we compare the interaction samples generated with LLMs to those collected in four previous behavior change studies. In these former studies, an RL agent chose different persuasive strategies for quitting smoking \cite{albers2023persuadingtoprep}, proposed different activities for quitting smoking \cite{albers2025reinforcement} and mental well-being \cite{degroot2025responsibleUNDER-REVIEW}, and decided whether to add human feedback when preparing for quitting smoking \cite{albers2025psychological}. For each of these studies, we assess how useful LLM-generated samples would be for two use cases. In the first use case, we want to learn effective actions to take in each state (i.e., a policy) that we can use as a warm start for a new application. So we determine how well an LLM-based optimal policy performs compared to one learned from the real behavioral data. In the second use case, we want to use the generated samples to gain insights into the rewards and state transitions of different actions. This could, for example, help to make design choices regarding the actions to use. Hence, we investigate how similar the LLM-based reward and transition functions are to those estimated from the four former studies' real behavioral samples.

\begin{figure}[h!]
    \centering
    \includegraphics[width=0.7\linewidth]{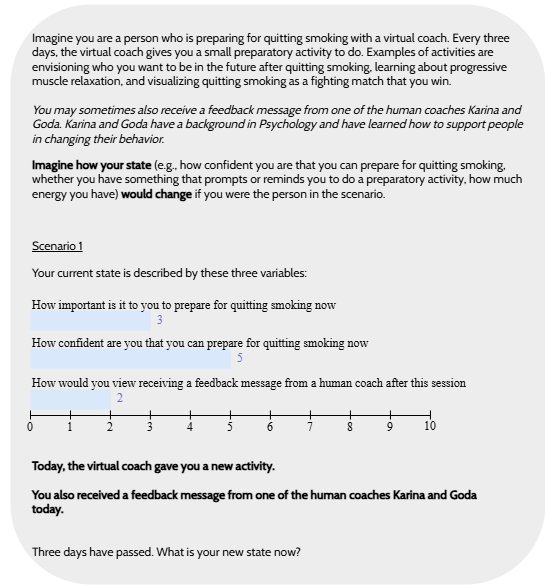}
    \caption{Illustration of the task of predicting next states for the behavior change study on deciding whether to add human feedback when preparing for quitting smoking. After receiving general information on the task and behavior change study, human raters were given scenarios describing an imaginary user's state (described by three variables illustrated with a bar chart) and a chosen action (receiving human feedback) for which to predict the next state. The information given to human raters is analogous to the one given to LLMs.}
    \label{fig:illustration_human_rating}
\end{figure}

We perform an extensive evaluation, including five different LLMs, a shorter and a more elaborate prompt version, ten paraphrases per prompt version, and an exploration of chain-of-thought prompting, few-shot prompting, and changing the model temperature. Our findings show that LLM-generated samples can lead to policies and reward and transition functions that outperform baselines. Asking human raters to predict rewards and next states (Fig.~\ref{fig:illustration_human_rating}) further shows that LLM-generated samples reach the performance of human-generated samples. The gap remaining between both LLM- and human-generated samples on the one hand and real behavioral samples on the other hand, however, highlights the difficulty in accurately predicting human behavior without real behavioral data. Moreover, we find based on an analysis of shorter and longer prompt versions, chain-of-thought prompting, and few-shot prompting that the relative effectiveness of these prompting strategies largely depends on the study, LLM, and use case. The only exception is few-shot prompting, which generally leads to improvements for at least one of the use cases. We provide recommendations for how LLM-generated samples can be useful in practice.

\section*{Related work}

\subsection*{Tailoring digital health applications for behavior change}

While meta-analyses show that digital health applications are effective for areas such as quitting smoking \cite{fang2023effectiveness,li2024efficacy}, increasing mental well-being \cite{groot2023effectiveness}, and reducing physical inactivity and unhealthy eating \cite{singh2024systematic}, users commonly do not adhere to these applications over a longer period of time or stop using them entirely after a while \cite{kelders2020concept, jabir2024attrition,shams2023adherence}. Instead of developing one-size-fits-all interventions, tailoring interventions to the distinct needs of specific (groups of) users could potentially improve adherence and reduce dropout \cite{singh2024systematic}. As such, there are a variety of approaches to adapt \textit{what} the applications offer, \textit{when}, \textit{how}, and \textit{with whom}. For example, \citet{dierikx2024collab} and \citet{gasparetti2020personalized} adapted daily step goals, \citet{wang2021optimizing} and \citet{trella2023reward} tailored the timing of notifications for physical activity and oral self-care, \citet{bertolotti2020different} and \citet{di2019applying} chose differently framed messages for healthy eating, and \citet{piette2023automatically} and \citet{forman2023using} learned when to add human support in digital applications for reducing opioid analgesic misuse risk and supporting weight loss. Group-level tailoring is thereby achieved by accounting for dynamic factors or states (e.g., available time, sleep quality, motivation, self-efficacy, and mood \cite{dierikx2024collab}) as well as more stable user characteristics (e.g., personality \cite{halko2010personality}, culture \cite{oyibo2018susceptibility}, motivation type \cite{alqahtani2023motivation}, stage of behavior change \cite{Oyebode2023persuasive}). Tailoring to individuals, either alone (e.g., \cite{kaptein2015personalizing}) or combined with group-level tailoring (e.g., \cite{tomkins2021intelligentpooling}), is also common. 

\subsection*{Computational methods for tailoring to users}

Various computational methods have been used to tailor elements of digital health applications to users, including classification-based methods and algorithms comparing user behavior to norms and guidelines \cite{klooster2024systematic}. For example, \citet{hors2019opening} used a recommender system to choose motivational messages for quitting smoking, \citet{albers2023settingphysical} selected testimonials for becoming more physically active based on a linear regression model, and \citet{ranjbartabar2021you} chose relational cues for a conversational agent based on users' preferences for examples provided before the interaction. An especially promising approach is reinforcement learning (RL) due to its ability to 1) learn which intervention options are optimal by trying them out, a process called exploration, 2) account for current user states when choosing intervention options as well as the effects of the latter on future states and thus the effectiveness of future intervention options, and 3) adapt to changes in these effects arising in a live application as long as some level of exploration remains. As such, RL has been applied to domains such as oral self-care (e.g., \cite{trella2025deployed}), physical activity coaching (e.g., \cite{wang2021optimizing}), and smoking cessation (e.g., \cite{albers2022addressing}), with initial results highlighting the potential of these algorithms to make behavior change support more effective.

\subsection*{Designing reinforcement algorithms for health behavior change}

There is, however, not ``one" RL algorithm. For example, while \citet{albers2022addressing} choose a \textit{persuasive strategy} that maximizes the \textit{engagement} with the intervention \textit{every few days}, \citet{wang2021optimizing} determine \textit{multiple times per day} whether to send a running \textit{notification} with the goal of maximizing a person's \textit{behavior change}. More generally, designers of RL algorithms need to make a variety of design choices \cite{trella2022designing}. This includes the decision times (e.g., send a message every day vs. every hour), the variables to include in the state description, the possible intervention options, the reward to be maximized, and the algorithm itself (e.g., whether to account only for current or also for future user states). Although behavior change theories and domain experts can provide guidance in making some of these choices (e.g. \cite{potter2023integrating}), they often cannot specify human behavior dynamics at a level of detail that is granular enough to predict the long-term effects of similar alternatives \cite{klasnja2015microrandomized}. Therefore, since testing many algorithm variants with real users is both costly and undesirable due to the real, negative consequences that sub-optimal intervention options can have for people, simulations are often used to evaluate different algorithm variants \cite{weimannunleashing2024,greene2023taking}. However, creating such simulations at a sufficient level of detail requires data on related human behavior, either for creating the model (e.g., \cite{wang2021optimizing,li2024dyadicreinforcementlearning,mintz2020nonstationary}) or for validating a literature-based model (e.g., \cite{jabeen2021healing}). Such data is often not publicly available \cite{weimannunleashing2024}.

\subsection*{LLMs taking the role of human participants}

The alternative for both testing different algorithm variants and pretraining new algorithms before deployment that we explore in this paper is generating human interaction samples with LLMs. Previous studies have provided promising results for using LLMs to take the role of human participants in various settings. For example, \citet{bavaresco2025llmsinsteadofhuman} show that LLMs can, in some cases, be adequate annotators for NLP tasks, which could speed up the (initial) evaluation of NLP models even if the LLMs should first be validated against human annotations for a particular task. Similarly, \citet{hamalainen2023evaluatingLLMs} show that LLMs can generate believable descriptions about experiencing video games as art, which could be useful for piloting new studies. And \citet{brugge2024large} use LLMs to simulate patients for medical students' training, giving medical students the opportunity to easily and cheaply prepare for medical conversations with a large number and a variety of case scenarios. There are inherent limitations to applying LLMs in these settings. For example, LLM-generated responses might contain factual errors or show less diversity than actual human responses \cite{hamalainen2023evaluatingLLMs}. They might also be less representative of people with a lower socioeconomic position or racial minorities \cite{kim2024aiaugmentedsurveysleveraginglarge} or have gender biases (e.g., \cite{bajaj2024evaluating}). However, the goal in each of these cases is not to entirely replace human participants, but to make better use of their time by identifying more promising NLP models, study questions, or conversation techniques in advance. These more promising variants can then be validated with real behavioral data as a next step. Examining how effective LLMs are for generating interaction samples to gain initial insights on RL models for digital behavior change applications is the goal of this paper.

To summarize, research has shown that while RL is a promising approach to create digital applications for behavior change that are effectively tailored to users and their states, it is in practice difficult to make all required design choices in an informed way when no human data is available. Since previous studies indicate that LLMs can potentially take the role of human participants, we investigate in this paper whether LLMs can also generate samples of humans interacting with digital applications for behavior change that are useful for designing RL approaches.

\section*{Experimental setup}

\begin{figure}[h!]
    \centering
    \includegraphics[width=\linewidth]{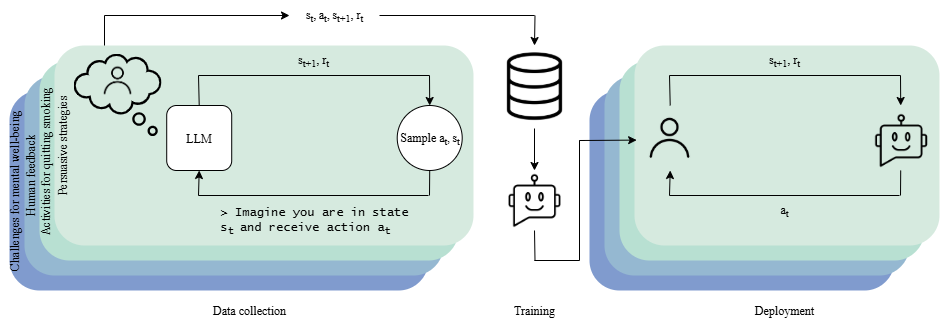}
    \caption{Overview of our offline RL approach applied to four different behavior change studies. Data collection: In the absence of real behavioral data, data is collected by prompting an LLM to imagine a user in state $s_t$ receiving action $a_t$. Training: The data is used to train an RL agent. Deployment: The trained agent is tested in a simulation of real users created using previously collected real behavioral data.}
    \label{fig:overview}
\end{figure}

To examine whether LLMs can generate useful samples for digital health behavior change settings, we take an offline RL approach on four different behavior change studies (Fig.~\ref{fig:overview}). Specifically, we ask LLMs to imagine a user in a state $s_t$ receiving an action $a_t$ and predict for the user the reward $r_t$ and next state $s_{t+1}$ after receiving $a_t$. Let's take the behavior change study on giving human feedback illustrated in Fig.~\ref{fig:illustration_human_rating} as example. A user's state is described by three state features that could influence whether human feedback is effective in the sense that it increases people's effort spent on smoking cessation activities. This includes how important preparing to quit is to the user, how confident they are to do it, and how much they would appreciate getting human feedback. So an LLM would, for example, be asked to imagine being a user with low perceived importance, high confidence and low feedback appreciation ($s_t$) who gets human feedback ($a_t$). For this user, the LLM would then predict the effort spent on a smoking cessation activity ($r_t$) and the next values for the three state features ($s_{t+1}$). 

All generated samples of the form $\langle s_t, r_t, a_t, s_{t+1}\rangle$ are used to learn a policy for an RL agent that describes which action to optimally take in which state. For the first use case, this policy is then evaluated in a simulation of real users that is created using real behavioral data from the behavior change study. For the second use case, we also look specifically at the reward and transition functions estimated from the generated samples in the process of learning a policy. These reward and transition functions, together also referred to as dynamics, describe the rewards and state transitions for all state-action combinations. We repeat this process for different settings: 1) using chain-of-thought prompting, 2) changing the temperature of the LLM, and 3) using few-shot prompting. Moreover, we compare against samples generated by asking human raters to make predictions analogous to the ones made by LLMs.

\subsection*{Behavior change studies}

We build upon four previously conducted behavior change studies. Table~\ref{tab:overview_studies} describes the state space, action space, reward, number of real behavioral samples, and the policy evaluation criterion for each of them. More information on the studies can be found elsewhere \cite{albers2023persuadingtoprep,albers2025reinforcement,albers2025psychological,degroot2025responsibleUNDER-REVIEW}.

\afterpage{%
\begin{landscape}
\begin{table}
  \caption{Overview of the four behavior change studies. We provide for each study information on the state space, action space, reward, number of collected real behavioral samples, and the policy evaluation criterion. One sample takes the form $\langle s_t, a_t, r_t, s_{t+1} \rangle$, where $s_t$ is a user's state at time $t$, $a_t$ is an action taken in state $s_t$, $r_t$ is the reward obtained for taking action $a_t$ in state $s_t$, and $s_{t+1}$ is the user's next state after action $a_t$ was taken. The evaluation criterion describes the objective according to which policies for the studies are evaluated in simulations in the original papers.}
  \label{tab:overview_studies}
  \begin{tabular}{p{6.0cm}p{4.1cm}p{3.1cm}cp{3.6cm}}
  \toprule
  \multicolumn{1}{c}{\bf State space} &\multicolumn{1}{c}{\bf Action space} &\multicolumn{1}{c}{\bf Reward} &\multicolumn{1}{c}{\bf \# of samples}&\multicolumn{1}{c}{\bf Evaluation criterion}\\ 
 \midrule
 \multicolumn{3}{l}{\textsc{Study 1: Persuasive strategies for quitting smoking}}\\
 Defined by 3 binary features: 1) whether users feel like they want to do
an activity, 2) whether they have things that prompt or remind
them to do an activity, and 3) whether they feel like they needed to
do an activity.& 5 persuasive strategies: 1) consensus, 2) commitment, 3) authority, 4) action planning, and 5) no persuasion. & Effort spent on preparatory activities for quitting smoking, rated on a scale from 0 to 10 and subsequently mapped to the interval $[-1, 1]$ &2366 &Reward (i.e., effort spent on preparatory activities for quitting smoking mapped to the interval [-1, 1])\\
\\
\multicolumn{3}{l}{\textsc{Study 2: Activities for quitting smoking}}\\
Defined by 9 features. The first 3 are binary and inquired from users: 1) belief that self-efficacy is useful for quitting, 2) belief that having the mindset that physical activity helps to quit smoking is useful, and 3) energy. The other 6 are deterministic and describe the degrees to which users have built 6 expert competencies for quitting smoking (e.g., practical skills, clear future identity, motivation). & 53 activities: 44 preparatory and 9 persuasive activities for quitting smoking. The 44 preparatory activities are thereby clustered into 5 actions for the prediction of the effort and the next nondeterministic state feature values, leading to a total of $5 + 9 = 14$ actions for the estimation of the learned parts of the dynamics.& Increase in expert competencies, which depends on the effort spent on activities and the activities' contributions to the competencies &1710 &Fraction of expert competencies built by the simulated users\\
\\
\multicolumn{3}{l}{\textsc{Study 3: Human feedback for quitting smoking}}\\
Defined by 3 features: 1) perceived importance of preparing to quit (3 values), 2) self-efficacy for preparing to quit (2 values), and 3) appreciation of receiving human feedback (2 values). & 2 actions: 1) not giving and 2) giving human feedback. & Effort spent on preparatory activities for quitting smoking, rated on a scale from 0 to 10 and subsequently mapped to the interval $[-1, 1]$ &2326 &Reward (i.e., effort spent on preparatory activities for quitting smoking mapped to the interval [-1, 1])\\
\\
\multicolumn{3}{l}{\textsc{Study 4: Coping challenges for mental well-being}}\\
Defined by 6 features. The first two influence the challenge completion: 1) tiredness (4 values), and 2) having completed a challenge the previous day (2 values). The other four track the number of challenges completed per coping strategy and influence only the diversity cost.& 4 actions: challenges corresponding to 4 different coping strategies. & Challenge completion (no/yes), with a cost to promote a diversity of coping strategies among the completed challenges & 5199 &Degree to which the simulated users have completed a diverse set of coping challenges, measured by the fraction of completing 4 challenges for each coping strategy\\ 
\bottomrule
  \end{tabular}
\end{table}
\end{landscape}
}

\subsection*{Models}

We evaluated five diverse open-weight models: Llama-3.2-3B-Instruct, Llama-3.1-8B-Instruct, Llama-3.3-70B-Instruct \cite{metaAI2024llama3blog}, Mistral-7B-Instruct-v0.3 \cite{mistralAI2023mistral7b}, and Command-R-v01 \cite{cohere2024commandR}. Using Llama models of various sizes thereby allows us to see if and how model size impacts performance. We focus on open-weight models because of both cost and reproducibility concerns (e.g., \cite{dietrich2025performance}). Since our goal was to test an out-of-the-box approach to generating samples, we used these models with a seed for reproducibility and otherwise largely the default parameter settings (Table~\ref{tab:llm_parameter_settings} in the Appendix). To test the influence of the temperature parameter, we conducted additional experiments with temperatures of 0.1 and 0.9.

\subsection*{Prompts}

\begin{figure}
    \centering
    \includegraphics[width=\linewidth]{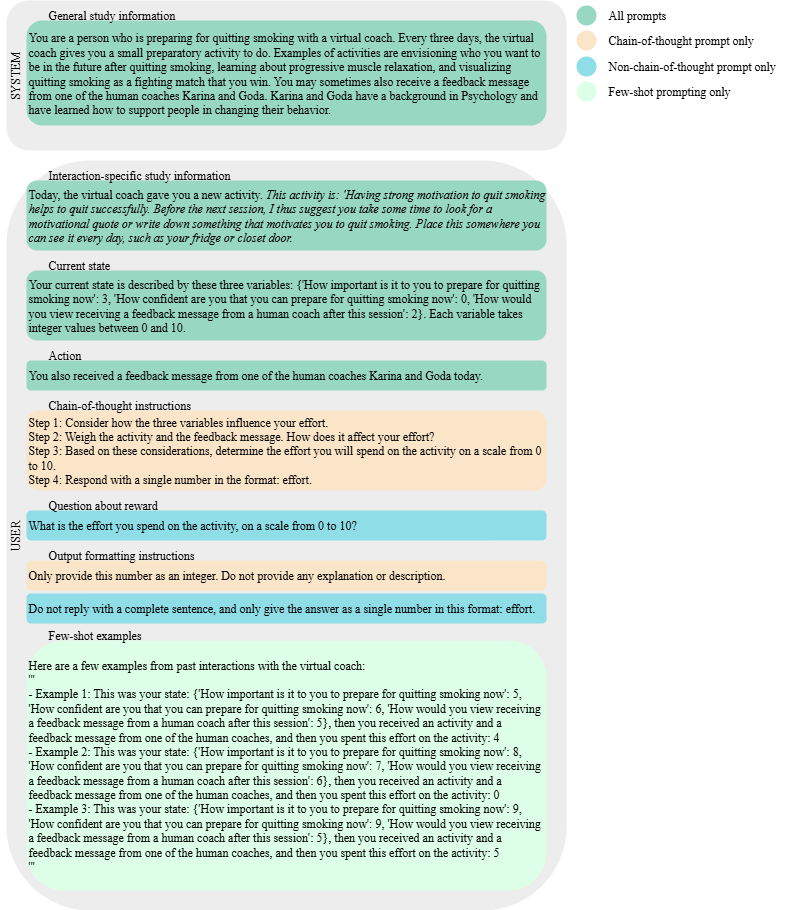}
    \caption{Example prompt for the reward for study 3 on human feedback for quitting smoking. The part in italics is only shown for the extensive prompt version.}
    \label{fig:prompt_example}
\end{figure}

For each behavior change study, we wrote a prompt that describes the study from the perspective of users, consisting of a description of 1) the study, 2) the user's current state, 3) the chosen action, and 4) a question about either the reward or the next state, together with output formatting instructions (Fig.~\ref{fig:prompt_example}). We created separate prompts for the reward and next state due to LLM output formatting difficulties when asking for both together. The Appendix shows the prompt templates for both the reward and the next state for all studies.

Since small changes to the prompt can lead to drastic changes in model outputs \cite{mizrahi2024state,sclar2024quantifying}, we performed all our experiments with 10 prompt variants. We created these by asking ChatGPT to rephrase the initial prompt while keeping the same meaning and subsequently correcting any inaccuracies introduced in this process (e.g., changed state variable names). Moreover, for studies 1--3, we created an extensive prompt version by providing more information on the content users received from the behavior change application (e.g., the full text of the preparatory activity assigned to the user or of the persuasive message sent to the user) if this content would be known before conducting a study. Note that we could not provide the full text of human feedback messages in the extensive prompts for study 3, because these messages would not be available in advance. For study 4, due to the very short text of the challenges, we did not vary the prompt length. Furthermore, to explore the effect of chain-of-thought prompting \cite{wei2022chain}, we also created a chain-of-thought version of the prompts (Fig.~\ref{fig:prompt_example}), again in 10 variations. We obtained these chain-of-thought prompts by asking ChatGPT to turn the base prompt into a chain-of-thought prompt and correcting the resulting prompt for introduced errors. Lastly, to examine the effect of few-shot prompting \cite{brown2020language}, we added between two and ten randomly chosen real behavioral samples with the same action as the one we were currently generating samples for (Fig.~\ref{fig:prompt_example}).

\subsection*{Human-generated samples: Asking human raters to generate samples}

If we find a difference in performance between the LLM-generated and the real behavioral samples, we do not know if this difference is due to LLMs being an unsuitable approach for predicting behavior or simply due to an inherent gap between predicting and actually performing a behavior. To better understand how good LLMs are at predicting behavior, we hence also asked human raters to generate samples. Specifically, we asked people to provide reward and next state predictions for 32 state-action combinations from one of the four studies (Fig.~\ref{fig:human_survey_screenshot} in the Appendix). We provided the information contained in the base prompt for studies 1--3 and the single prompt length we have for study 4. Participants were recruited via convenience sampling as well as through the student participant pool at the Tilburg School of Humanities and Digital Sciences and allocated to the four studies such that studies with more actions had a proportionally higher chance of being chosen. All participants had to give digital informed consent before providing predictions. Participants were included as long as they passed at least one attention check question, and data from participants who failed more than one attention check question was removed. Overall, we obtained between about 76 (study 2, 14 actions) and 123 (study 3, 2 actions) samples per action for the four studies. The study was reviewed and approved by the Research Ethics and Data Management Committee of the Tilburg School of Humanities and Digital Sciences, belonging to Tilburg University, with code REDC 2025.37. Before data collection, the study was preregistered in the Open Science Framework (OSF) \cite{albers_krahmer_2025}. Analogously to LLM-generated samples, we call the resulting samples human-generated samples. This is not to be confused with the real behavioral samples, which, while also stemming from humans, are not generated in the same sense.

\section*{Results}

\subsection*{How well do the learned policies perform?}

\paragraph{Setup}

In the first use case, we want to learn an effective policy that we can use as a warm start for a new application. To assess how well the policies learned based on samples from the different models perform, we used simulations analogous to the ones performed in the original papers \cite{albers2023persuadingtoprep, albers2025reinforcement,albers2025psychological,degroot2025responsibleUNDER-REVIEW}. As described in Table~\ref{tab:overview_studies}, each study thereby has its own evaluation criterion. For studies 1 and 3, the simulations keep track of the effort people spend on preparatory activities for quitting smoking over time. For study 2, the goal is to allow all simulated users to fully build six competencies for quitting smoking. For study 4, we want users to complete a diverse set of coping challenges, measured by their progress in completing four challenges for each coping strategy. We show means and 95\% credible intervals based on the 10 prompt variants. As comparisons served an optimal policy $\pi^*$, a worst policy $\pi^-$, and the optimal policy $\pi^H$ computed based on the samples from the human evaluation. For studies 2 and 4, we additionally show the performance of a policy that chooses random activities or coping challenges ($\pi^r$). Moreover, since the dynamics for studies 2 and 4 consist of both deterministic and learned components, we also consider a policy that only considers deterministic and thus no learned dynamics when choosing an optimal activity. Analyses are based on 500 generated samples per action. We chose this number of samples both because it falls roughly in the middle of the number of real samples we have from the four behavior change studies (Table~\ref{tab:overview_studies}) and because analyses for the second use case below show that there are already hardly any changes beyond 200 generated samples per action (Fig.~\ref{fig:l1_error_reward} and Fig.~\ref{fig:l1_error_transitions}).

\paragraph{Results}

\begin{figure}[h!]
 \centering
  \includegraphics[width = \textwidth]{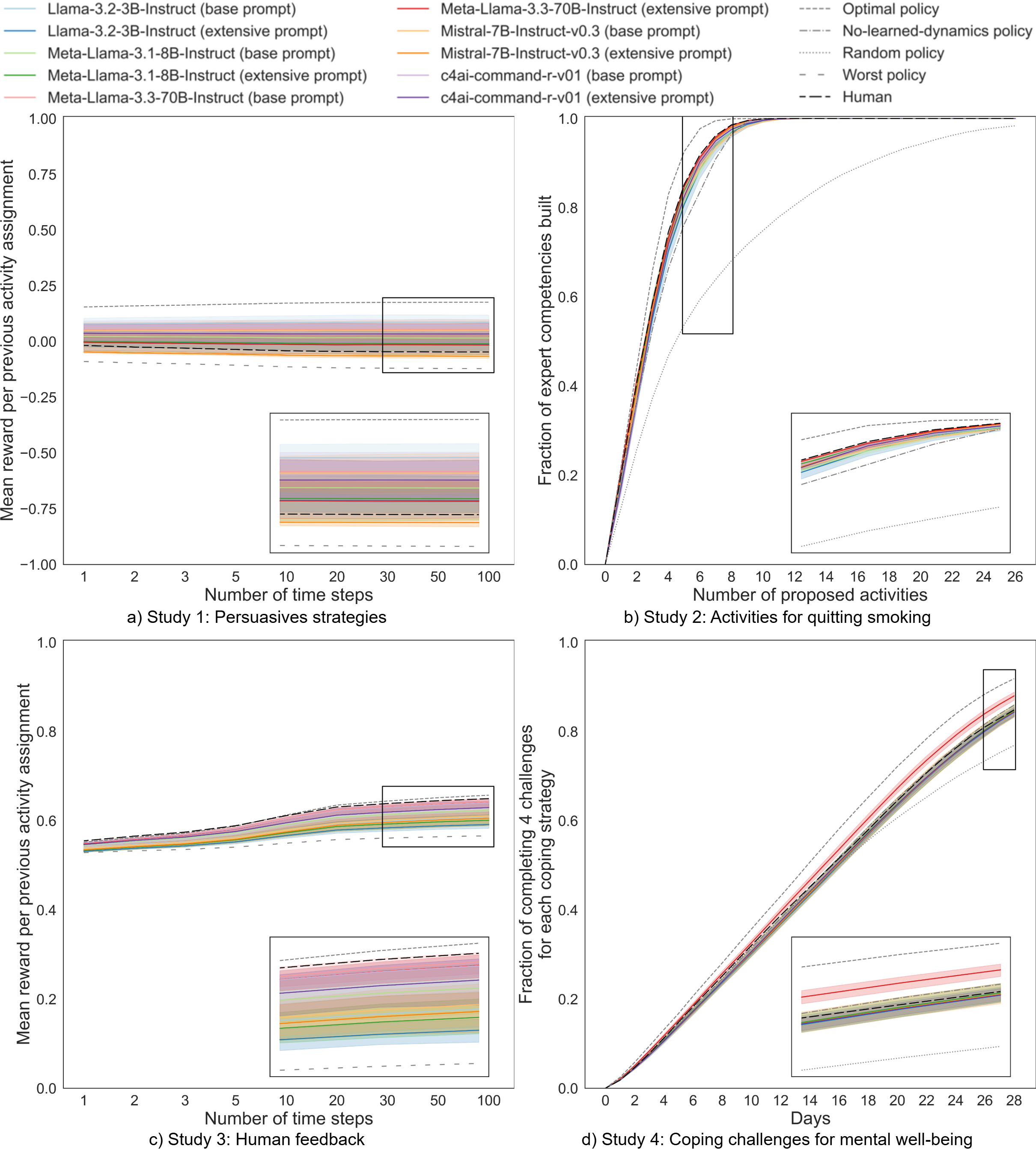}
  \caption{\small Simulated performance of policies learned from samples generated by different LLMs compared to policies learned from the real behavioral samples as well as human-generated samples. For each of the four behavior change studies described in Table~\ref{tab:overview_studies}, the policies are evaluated on the study-specific evaluation criterion (y-axis) over time (x-axis). The \texttt{optimal policy $\pi^*$}, \texttt{no-learned-dynamics policy}, and the \texttt{worst policy $\pi^-$} are learned based on the real behavioral data. The \texttt{human policy $\pi^H$} is learned based on the human-generated samples.}
  \label{fig:policy_performance_simulated}
\end{figure}

Fig.~\ref{fig:policy_performance_simulated} summarizes the results by showing for each behavior change study how different LLM-based policies perform on the study-specific evaluation criterion over time. For example, the pale red line in Fig.~\ref{fig:policy_performance_simulated}c represents the performance of the policy learned from the samples generated by Llama-3.3-70B with the basic prompt for study 3. Llama-3.3-70B is the best-performing LLM for that study, almost reaching the performance of $\pi^*$. Generally, one can see that while all LLM-based policies outperform the lowest baselines (i.e., $\pi^-$ for study 1 and 3 and $\pi^r$ for study 2 and 4) and for study 2 also the policy that does not consider any learned dynamics, only one model-study combination reaches the performance of the optimal policy, namely, Llama-3.3-70B for study 3. Moreover, for study 4, Llama-3.3-70B is also the only model that performs better than a policy that does not take any learned dynamics into account. Compared to human performance, we generally see that the best LLMs reach human performance. For studies 1 and 4, some LLMs numerically even outperform $\pi^H$. However, both human and LLM performance is lower for these than for the other studies in the sense that the performance is further away from that of $\pi^*$ (study 1) or comparable to a policy that does not consider any learned dynamics (study 4). No prompt type (i.e., basic vs. extensive) consistently performs best. For example, the basic prompt tends to slightly outperform the extensive prompt for studies 1 and 3, whereas the reverse is the case for study 2. Regarding models themselves, there is also no overall best model. However, Llama-3.3-70B tends to be among the best models.

\subsection*{How good are the learned reward and transition functions?}

\paragraph{Setup}

In the second use case, we want to use the generated samples to gain insights into the rewards and state transitions (i.e., dynamics) of different actions. Therefore, we computed how close the dynamics estimated based on the LLM samples are to the dynamics estimated from the real behavioral data. For this, we computed mean absolute errors (i.e., $L_1$-errors) between the real and generated reward and transition functions for different numbers of samples per action. To contextualize the performance, we computed the human baseline as well as simple data-independent baselines. The latter are assuming that people spend the mean reward for each state-action combination for the reward, and assuming that all transitions are equally likely or that people stay in their state for the transitions. Moreover, we show the results of drawing the same number of samples from the real behavioral samples (i.e., from an oracle). We provide means and 95\% credible intervals based on the 10 prompt variants for the LLM-generated samples and based on 10 random draws for the real behavioral samples.

\paragraph{Results}

\begin{figure}[h!]
 \centering
 \includegraphics[width=\textwidth]{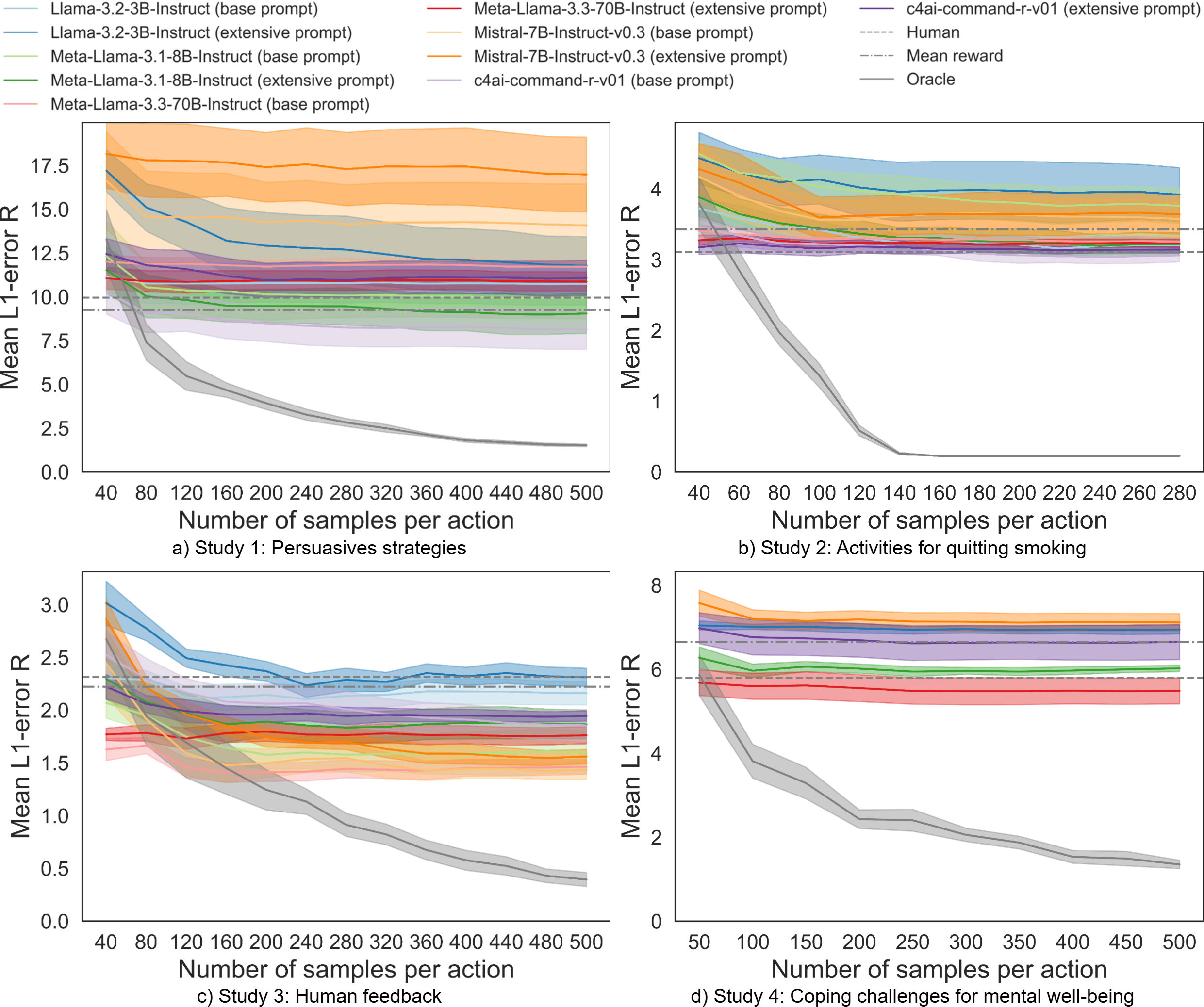}
  \caption{\small Mean $L_1$-error and 95\% credible interval between the rewards estimated from samples generated by different LLMs and those estimated from the real behavioral samples for different numbers of samples per action. As comparisons serves assuming that people spend the mean reward for each state-action combination (\texttt{Mean reward}) as well as drawing samples from the real behavioral samples (\texttt{Oracle}). Moreover, we computed the reward function from the human-generated samples (\texttt{Human}). Means are shown over the 10 different prompt variants for LLM-generated samples and over 10 random draws for the real behavioral samples.}
  \label{fig:l1_error_reward}
\end{figure}

\begin{figure}[h!]
 \centering
  \includegraphics[width=\textwidth]{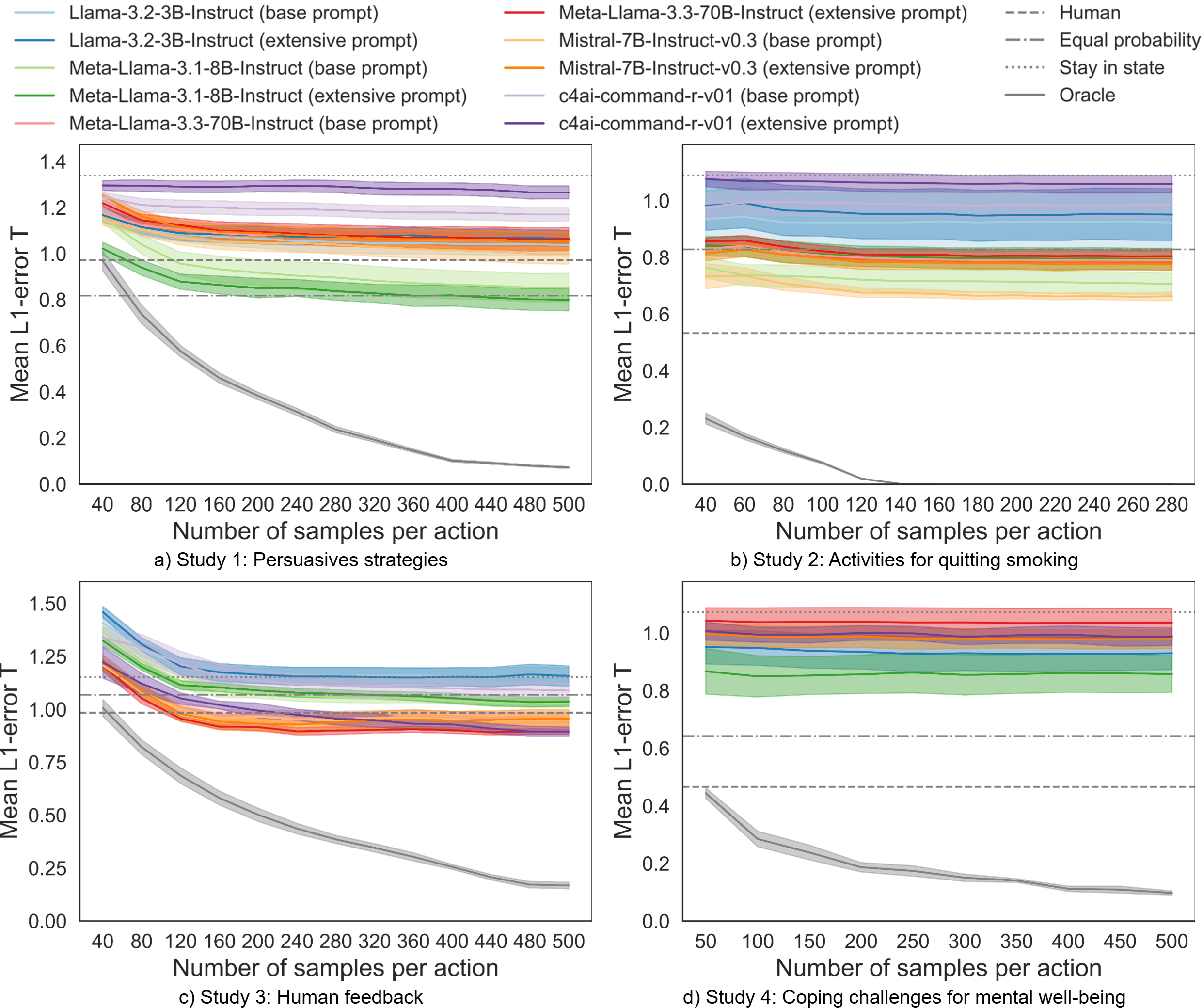}
  \caption{\small Mean $L_1$-error and 95\% credible interval between the transition functions estimated from samples generated by different LLMs and those estimated from the real behavioral samples for different numbers of samples per action. As comparisons serves assuming that all transitions are equally likely (\texttt{Equal probability}), assuming that people stay in their state (\texttt{Stay in state}), and drawing samples from the real behavioral samples (\texttt{Oracle}). Moreover, we compute the transition functions based on the human-generated samples (\texttt{Human}). Means are shown over the 10 different prompt variants for LLM-generated samples and over 10 random draws for the real behavioral samples.}
  \label{fig:l1_error_transitions}
\end{figure}

Fig.~\ref{fig:l1_error_reward} and Fig.~\ref{fig:l1_error_transitions} show for each behavior change study the mean $L_1$-errors for the reward and transition functions, respectively, estimated from samples generated by different LLMs. For example, the red line in Fig.~\ref{fig:l1_error_reward}d represents the $L_1$-error between the reward function estimated from the real behavioral samples from study 4 and the one estimated from the samples generated by Llama-3.3-70B. This reward function by Llama-3.3-70B is more accurate than assuming that people always spend the mean reward (\texttt{Mean reward}) and the reward function estimated from the human-generated samples (\texttt{Human}), but is less accurate than estimating the reward function based on a subset of the real behavioral samples (\texttt{Oracle}). Similarly, the red line in Fig.~\ref{fig:l1_error_transitions}d represents the $L_1$-error between the transition function estimated from the real behavioral samples from study 4 and the one estimated from the samples generated by Llama-3.3-70B. Here, Llama-3.3-70B is one of the worse performing LLMs, performing worse than all comparisons other than assuming that people stay in their state (\texttt{Stay in state}). 

Generally, one can see that the best LLMs tend to be better than the simple data-independent baselines for both rewards (Fig.~\ref{fig:l1_error_reward}) and transitions (Fig.~\ref{fig:l1_error_transitions}). This ranges from almost all LLMs outperforming the simple baselines for study 3 (Fig.~\ref{fig:l1_error_reward}c and Fig.~\ref{fig:l1_error_transitions}c) to only a single LLM for study 1 (Fig.~\ref{fig:l1_error_reward}a and Fig.~\ref{fig:l1_error_transitions}a). Interestingly, study 1 also seems to be harder for humans in that human performance is worse than the mean reward baseline for the rewards and worse than assuming equal transition probabilities for the transitions. For all other studies, human performance reaches or exceeds the performance of the simple baselines. Generally, comparing LLM to human performance again shows that the best LLMs tend to be similar to human performance. For the rewards and transitions for study 3 (Fig.~\ref{fig:l1_error_reward}c and Fig.~\ref{fig:l1_error_transitions}c), several LLMs even perform better than the human baseline. On the other hand, none of the LLMs reach human performance for the transitions for studies 2 and 4 (Fig.~\ref{fig:l1_error_transitions}b and Fig.~\ref{fig:l1_error_transitions}d). In both of these cases, human performance, while better than the simple baselines, does not reach the performance of the oracle with even very few samples, suggesting that actual and imagined transitions are quite different for those two studies. This observation that actual and imagined dynamics differ also holds across the studies and types of dynamics, as between 40 and 200 real behavioral samples per action are enough for the oracle to outperform any LLM as well as human raters. Regarding the types of LLMs and prompts, we again find that while no LLM is always the best, Llama-3.3-70B tends to be among the best. And whether the base prompt outperforms the extensive one depends on the study and model, but the reverse generally does not happen. Lastly, note that the $L_1$-errors for the LLMs plateau rather quickly, with more than around 200 samples per action not providing much additional benefit.

\subsection*{Does chain-of-thought prompting help?}

\paragraph{Setup}

To see whether chain-of-thought prompting helps for the two use cases, we repeated our policy simulations and analysis of the accuracy of learned dynamics with and without chain-of-thought prompting for a subset of the LLMs, making sure to include Llama-3.3-70B as one of the better models, as well as two smaller models for comparison. We again show means and 95\% credible intervals based on the 10 prompt variants for the generated samples and 10 random draws for the real behavioral samples. Since the earlier analyses showed that the accuracy of the dynamics estimates plateaus around 200 samples per action, we used 300 generated samples per action for the policy simulations to reduce the comparatively longer time needed to generate samples based on chain-of-thought prompting. Table~\ref{tab:chain-of-thought_reasonings_example_reward} and Table~\ref{tab:chain-of-thought_reasonings_example_transition} in the Appendix show examples of the LLM output for chain-of-thought prompts when removing the output formatting restrictions from the prompt.

\paragraph{Results}

Fig.~\ref{fig:policy_performance_simulated_cot}, Fig.~\ref{fig:l1_error_rewards_cot}, and Fig.~\ref{fig:l1_error_transitions_cot} in the Appendix show the results for the policy performance, accuracy of the reward function, and the accuracy of the transition function, respectively. For instance, the light and dark green lines in Fig.~\ref{fig:l1_error_rewards_cot}a represent the accuracies of the reward functions estimated based on samples generated by Llama-3.1-8B without and with chain-of-thought prompting for study 1. The reward function estimated from samples generated without chain-of-thought prompting (light green) thereby has a lower $L_1$-error. Overall, we find that whether chain-of-thought prompting helps depends again on both the study and the LLM. For example, while chain-of-thought prompting tends to make the reward estimation worse for study 3 (Fig.~\ref{fig:l1_error_rewards_cot}c in the Appendix), it makes it better for study 1 with the exception of Llama-3.1-8B (Fig.~\ref{fig:l1_error_rewards_cot}a). For the transitions, there is an improvement for study 4, a deterioration for study 1, and generally no difference for studies 2 and 3 except for one model each where the $L_1$-error gets worse (Fig.~\ref{fig:l1_error_transitions_cot} in the Appendix). And the policies learned with chain-of-thought prompting perform worse for Llama-3.3-70B for study 3, better for Llama-3.3-70B with the extensive prompt for study 1 and Llama-3.1-8B for study 4, and similar for all other settings (Fig.~\ref{fig:policy_performance_simulated_cot} in the Appendix).

\subsection*{Does changing the temperature help?}

\paragraph{Setup}

To assess the impact of the temperature set for the LLMs, we compared our policy simulations and analysis of the accuracy of learned dynamics for temperatures of 0.1, 0.6, and 0.9. We only used Llama-3.3-70B as the overall best performing model from our initial analyses. We used both the basic and the extensive prompt, and again show means and 95\% credible intervals based on the 10 prompt variants for the generated samples and 10 random draws for the real behavioral samples. Results for the policy simulations are based on 500 generated samples per action.

\paragraph{Results}

Fig.~\ref{fig:policy_performance_simulated_temp}, Fig.~\ref{fig:l1_error_rewards_temp}, and Fig.~\ref{fig:l1_error_transitions_temp} in the Appendix summarize the results for the four behavior change studies with respect to policy performance, the $L_1$-error for the reward function, and the $L_1$-error for the transition function, respectively. For instance, the three blue lines in Fig.~\ref{fig:policy_performance_simulated_temp}c show the performance of the policies learned based on samples generated by Llama-3.3-70B with the base prompt with different temperatures for study 3. While the means of the three policies differ slightly, the 95\% credible intervals overlap, suggesting that there is no clear difference in performance. Overall, we find that changing the temperature generally does not affect the policy performance or the accuracy of the dynamics. Only for the extensive prompt for study 1 does a higher temperature lead to a slightly better policy performance, albeit still one that is worse than for the basic prompt (Fig.~\ref{fig:policy_performance_simulated_temp}a).

\subsection*{Does few-shot prompting help?}

\paragraph{Setup}

To see whether few-shot prompting leads to better performing policies and more accurately estimated dynamics, we repeated our analyses one more time with between two and ten real behavioral samples added to the prompt for Llama-3.3-70B. These real samples were randomly drawn from all real samples with the same action that we were currently asking the LLM to make predictions for. Since we did not have ethical approval to transfer the real behavioral samples from study 4 to the server where the LLMs were run, we focused solely on the first three behavior change studies.

\paragraph{Results}

Fig.~\ref{fig:policy_performance_simulated_fewshot}, Fig.~\ref{fig:l1_error_rewards_fewshot}, and Fig.~\ref{fig:l1_error_transitions_fewshot} in the Appendix summarize the results of adding real behavioral samples to the prompt, this time for only three instead of four behavior change studies. For instance, Fig.~\ref{fig:l1_error_transitions_fewshot}b shows that adding any number of real behavioral samples to the prompt for study 2 (blue lines) leads to a more accurate transition function than using no real samples (red line). Generally, we observe that few-shot prompting is most beneficial for estimating the transition function, with clear improvements for studies 2 and 3 and slight improvements for five and ten examples for study 1 (Fig.~\ref{fig:l1_error_transitions_fewshot} in the Appendix). For the rewards, we find that few-shot prompting leads to clear improvements for study 1 as long as more than two examples are provided, no changes for study 1, and a clear deterioration for study 3 the more examples are given (Fig.~\ref{fig:l1_error_rewards_fewshot}). Effects on the policy performance are mixed as well, with no clear changes for study 1 and a slight deterioration for study 2 (Fig.~\ref{fig:policy_performance_simulated_fewshot}a and Fig.~\ref{fig:policy_performance_simulated_fewshot}b). For study 3, the zero-shot policy already performs near-optimal, which means it is difficult to perform better. Few-shot prompting with five and ten examples reaches that performance, with all other numbers of examples leading to worse performance (Fig.~\ref{fig:policy_performance_simulated_fewshot}c). 

\section*{Discussion}

Based on generating samples with five different open-weight LLMs for four different behavior change studies, we find that LLM-generated samples make it possible to learn policies that outperform worst and random baselines in human data-based simulations. Moreover, reward and transition functions estimated from the generated samples are generally closer to those estimated from the real behavioral samples than heuristics such as assuming people stay in their state or that all transitions are equally likely. This suggests that LLMs can be a useful tool for generating interaction samples for digital health behavior change settings.

\subsection*{Effect of different prompting strategies}

We find that chain-of-thought prompting neither consistently improves nor deteriorates the LLM-generated samples. One possible reason for this is that chain-of-thought prompting might be less effective for our models, in line with \citet{bajaj2024evaluating} finding that chain-of-thought prompting helped only for the smaller and older models they tested. Moreover, \citet{song2025burn} saw for multimodal LLMs that chain-of-thought prompting helped for some tasks such as ordering tasks but not others such as event relation inference. It could hence be that step-by-step reasoning is not necessarily helpful when it comes to predicting human responses to behavior change support strategies. In light of the findings by \citet{liu2025mindstepbystep} that step-by-step reasoning has sometimes similar negative effects for humans and LLMs, it would be interesting to see in the future whether the effects of step-by-step reasoning are also similar for humans and LLMs for our studies. Notably, our results are based on ten slightly different chain-of-thought prompts. It is thus possible that a single fine-tuned chain-of-thought prompt outperforms our original prompt. Our goal, however, was to see if a simple application of chain-of-thought prompting would be beneficial, as prompt fine-tuning is difficult in our envisioned setting where there is no real behavioral data. We find this not to be the case. Similarly, providing more information in the prompt or changing the temperature also has no consistent effect for our studies and models. Changing the temperature thereby has hardly any effect at all. This is good news for our envisioned setting where fine-tuning based on real behavioral data is difficult, as results should be relatively insensitive to the temperature setting. Lastly, providing between five and ten real examples per action tends to help for estimating the transition function, with mixed effects for the policy performance and reward estimation.

\subsection*{Comparing LLM and human rater predictions}

Both the performance of policies and accuracy of dynamics based on LLM-generated samples reach those of human raters who were asked to make predictions for people described in scenarios, suggesting that remaining differences between generated and real behavioral samples are due to a gap between envisioning and actually doing behavior rather than a limitation specific to using LLMs rather than human raters. Humans are not necessarily good at predicting their behavior due to biases such as the optimism bias, according to which people tend to be overly optimistic about themselves and their future \cite{weinstein1980unrealistic}. And we even went one step further by asking people to predict the behaviors of \textit{others} described in scenarios. Even though such scenario-based predicting of human behavior is often used in behavior change settings as an initial low-cost and low-risk evaluation step (e.g., \cite{deVries2016crowddesigned,ghantasala2023feasibility,tielman2019design}), it is evident from our observed gap between human-predicted and real behavioral samples that this is not an easy task for humans. Based on previous work finding that LLMs mirror certain human cognitive biases (e.g., temporal biases \cite{song2025burn} or heuristics for syllogistic inference \cite{bertolazzi2024systematic}), it would be interesting to investigate if this is also the case for our studies. Anecdotally, some human raters remarked after their participation that to better predict human behavior, they would need to know more about the people in the scenario. While one would not always know the exact target population in our envisioned setting, future work could examine whether adding user characteristics such as age or occupation sampled from a plausible range improves both LLM and human prediction performance. If such user characteristics should also be part of the RL algorithm, however, care needs to be taken of potential biases in LLM responses based on factors such as socioeconomic position (e.g., \cite{kim2024aiaugmentedsurveysleveraginglarge}), gender (e.g., \cite{bajaj2024evaluating}), or culture (e.g., \cite{naous2024havingBeer}).

\subsection*{Relative performance of LLMs}

Regarding the relative performance of the different models, no single LLM outperforms all other ones for all studies and metrics. Llama-3.3-70B is, however, usually one of the best-performing models. That no single LLM always performs best and the trend that larger Llama models tend to outperform their smaller counterparts is, for example, also what \citet{testoni2025mindgapbenchmarkingllm} observed for their estimation of uncertainty for medical question answering. Part of the reason why it is difficult to judge which models are best in our case is also that while the mean metrics of models often differ, their 95\% credible intervals based on different prompt variants overlap, thus not providing a clear indication that one model is better than another. This underlines the importance of evaluating LLMs with different prompt variants \cite{mizrahi2024state,sclar2024quantifying}. In our case, making slight changes to the prompt also led to sometimes drastic changes in the degree to which the models were following the required output format, making it necessary to add more logic for preprocessing the model outputs. This was done to avoid that observed differences are primarily due to different output formatting abilities rather than actual differences in the prediction of human responses. In case one is also interested in the former, the additional preprocessing could be removed. \citet{song2025burn}, for example, saw that one of the reasons for unsatisfactory performance of their multimodal LLMs in temporal grounding and reasoning tasks is poor output formatting.

\subsection*{Practical implications}

We set out this paper with the idea that in the absence of real behavioral data, generating interaction samples with LLMs could be useful for building RL models. Either to train a policy that can be used as warm start for a new application (use case 1), or to gain insights into the actual rewards and transitions that can, for example, be used to make design choices such as which actions to use (use case 2). Our findings show that LLMs can be a useful tool for generating samples in the absence of any real behavioral data. If a few real samples are available (e.g., five or ten per action), it seems to generally be useful to incorporate these into the prompt when aiming to estimate the transition function. And once between 40 and 200 real samples per action are available, these real samples clearly outperform any number of LLM-generated samples. Since we do not find consistent effects of providing more or less study information, providing more information beyond a minimum does not seem to be necessary, which is helpful if not all study details are known yet. Slight prompt paraphrases do, however, lead to different reward and transition functions, suggesting that using more than one might help to not get extreme results. 

Regarding model choice, Llama-3.3-70B generally performed best across studies and use cases. While it is possible that other LLMs, for example reasoning-based ones (e.g.,\cite{muennighoff2025s1simpletesttimescaling,deepseekai2025deepseekr1incentivizingreasoningcapability}), perform better than the five LLMs we tested in our proof-of-concept study, it is unlikely that the improvements are large. This is because our best LLMs tend to already reach the performance of human raters. Besides performance, a model choice should also take the costs of using LLMs into account. This includes the actual price of the required resources researchers or their institutes need to pay, but also externalized costs such as energy consumption, carbon emissions, and the use of clean water and minerals (e.g., \cite{luccioni2023estimating,luccioni2024power,luccioni2024environmental}), the effects of which tend to be disproportionally felt by already marginalized groups \cite{agyeman2003just}. It could also be that traditional simulation approaches, while requiring domain knowledge and programming expertise to build, are ultimately cheaper when all costs are considered. LLMs have also been explored as a way to reduce the amount of expertise required to build such simulations \cite{khanshan2024evaluation}, which while increasing the environmental cost would likely still result in fewer required LLM inferences than full LLM-based simulations.

\section*{Final remarks}

Overall, we show based on four diverse behavior change studies that an out-of-the-box approach to generating RL samples with LLMs can help to gain some insights into effective actions and the rewards and state transitions associated with them. Our answer to the effects of different prompting strategies is generally ``it depends" - effects often differ between studies and use cases. Only few-shot prompting seems to offer a more consistent advantage, but also primarily for estimating state transitions. We see our findings as supporting the idea of using LLM-generated samples as a tool when designing a new RL algorithm and encourage future work on how the LLM-generated samples can be effectively combined with real behavioral samples when they ultimately become available. Moreover, since predicting the effects of different actions chosen in a certain context forms the basis of tailoring in general, our approach also holds promise for tailoring approaches for digital health behavior change other than reinforcement learning.

\section*{Acknowledgments}

The authors acknowledge the help they received from Enrico Liscio in setting up the sample generation with LLMs.

\section*{Disclosure statement}

No potential conflict of interest was reported by the authors.

\section*{Author contributions}

NA and EK conceived the study. NA collected the data from LLMs and human raters, performed the analysis, created the figures, and wrote the main manuscript text under the supervision of EK. EG helped with the LLM data collection and the analysis for the study on coping challenges for mental well-being. LK and MH contributed the real behavioral data for the study on coping challenges for mental well-being. All authors contributed to reviewing the manuscript.

\section*{Funding}

This work is part of the project ``Technology Assisted Self-Management: Preventing Relapse and Crisis by the Severe Mentally Ill Themselves" with file number KICH1.GZ03.21.002 of the research program ``KIC - MISSIE Zorg in eigen leefomgeving 2021" which is (partly) financed by the Netherlands Organization for Scientific Research (NWO). This work further is part of the PROTECt ME project, funded by the Convergence, the alliance between Erasmus Medical Center Rotterdam, Erasmus University Rotterdam, and Delft University of Technology.

\section*{Data availability statement}

The real behavioral data from the first three behavior change studies is publicly available with the respective papers \cite{albers2023persuadingtoprep,albers2025reinforcement,albers2025psychological}. Access to the real behavioral data from the fourth behavior change study can be requested via the Grow It! team using a data request form that can be found in the \href{https://osf.io/q83hg?view_only=b691104ecc3d45ad8b48e1bd60ad7125}{Grow It! codebook}.
All other data and analysis code underlying this paper will be made available in a public repository upon paper acceptance.

\bibliography{bib}

\clearpage

\section*{Appendix}

\renewcommand{\thetable}{A\arabic{table}}
\renewcommand{\thefigure}{A\arabic{figure}}
\setcounter{figure}{0}
\setcounter{table}{0}

\subsection*{LLM parameter settings}

\begin{table}[h]
  \caption{Settings of LLM parameters used unless otherwise noted for our experiments with other temperatures. We also set a random seed for reproducibility. All other parameters were not specified, which means that default values were used.}
  \label{tab:llm_parameter_settings}
  \begin{tabular}{p{6.0cm}p{3.6cm}}
  \toprule
  \multicolumn{1}{c}{\bf Parameter} &\multicolumn{1}{c}{\bf Value}\\ 
 \midrule
 max\_tokens & 4096\\
 temperature & 0.6\\
 top\_p & 0.9\\
\bottomrule
  \end{tabular}
\end{table}

\subsection*{All prompt templates}

\begin{figure}
    \centering
    \includegraphics[width=\linewidth]{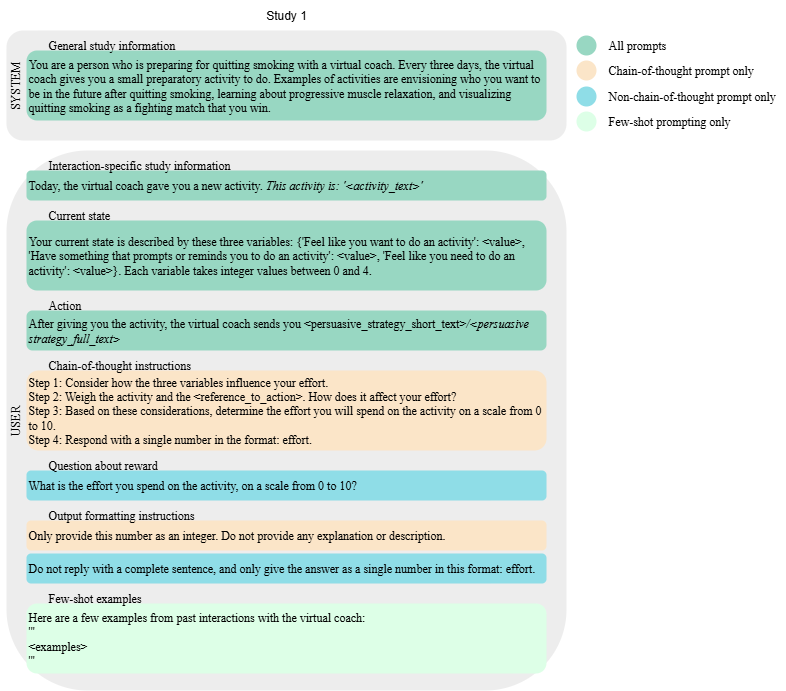}
    \caption{Prompt template for the reward for study 1 on persuasive strategies for quitting smoking. The part in italics is only shown for the extensive prompt version.}
    \label{fig:prompt_template_study1_reward}
\end{figure}

\begin{figure}
    \centering
    \includegraphics[width=\linewidth]{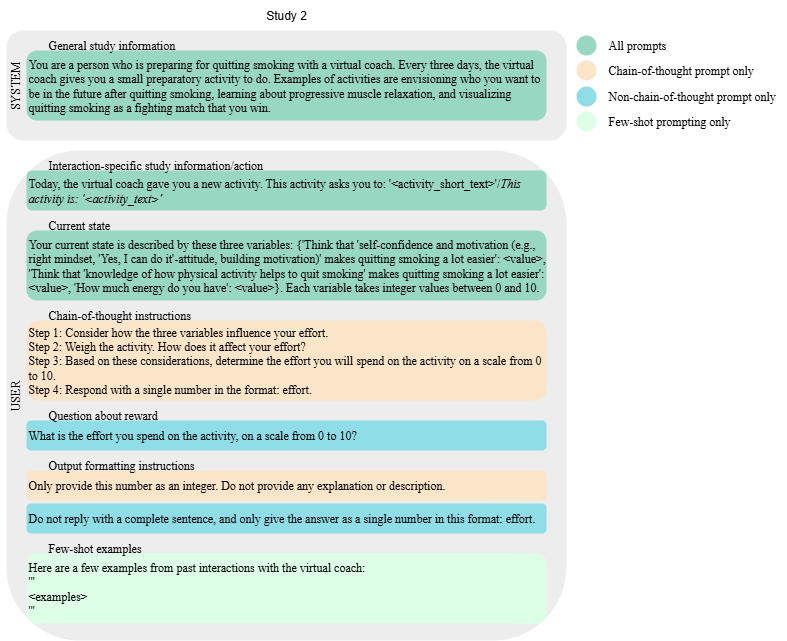}
    \caption{Prompt template for the reward for study 2 on activities for quitting smoking. The part in italics is only shown for the extensive prompt version.}
    \label{fig:prompt_template_study2_reward}
\end{figure}

\begin{figure}
    \centering
    \includegraphics[width=\linewidth]{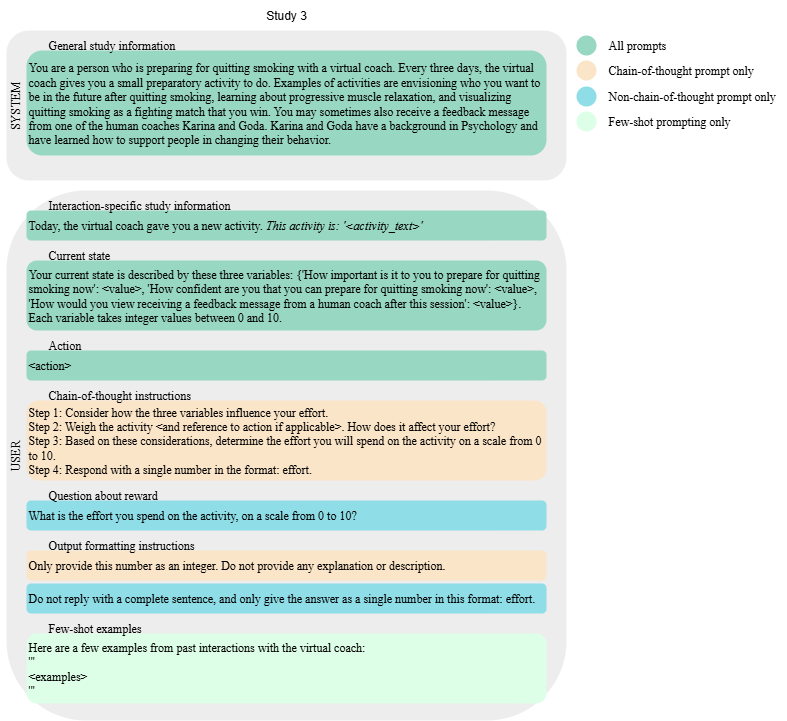}
    \caption{Prompt template for the reward for study 3 on human feedback for quitting smoking. The part in italics is only shown for the extensive prompt version.}
    \label{fig:prompt_template_study3_reward}
\end{figure}

\begin{figure}
    \centering
    \includegraphics[width=\linewidth]{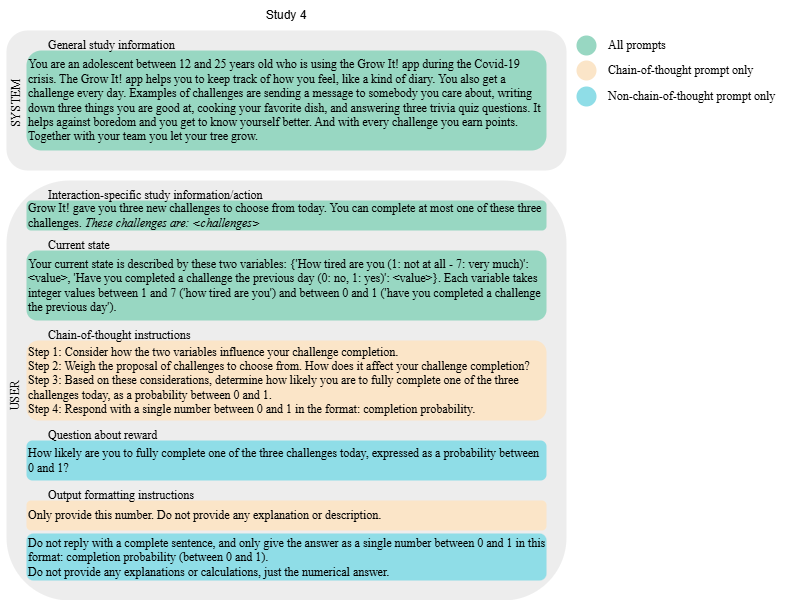}
    \caption{Prompt template for the reward for study 4 on coping challenges for mental well-being. The part in italics is only shown for the extensive prompt version, which is the only prompt version we use for this study due to the brevity of the challenge texts.}
    \label{fig:prompt_template_study4_reward}
\end{figure}

\begin{figure}
    \centering
    \includegraphics[width=\linewidth]{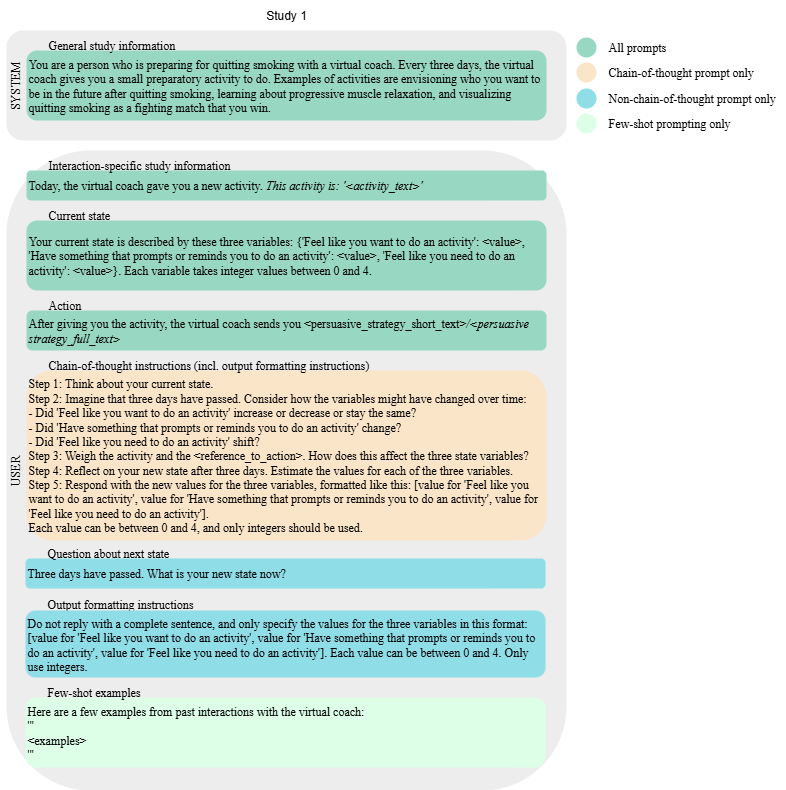}
    \caption{Prompt template for the next state for study 1 on persuasive strategies for quitting smoking. The part in italics is only shown for the extensive prompt version.}
    \label{fig:prompt_template_study1_nextstate}
\end{figure}

\begin{figure}
    \centering
    \includegraphics[width=\linewidth]{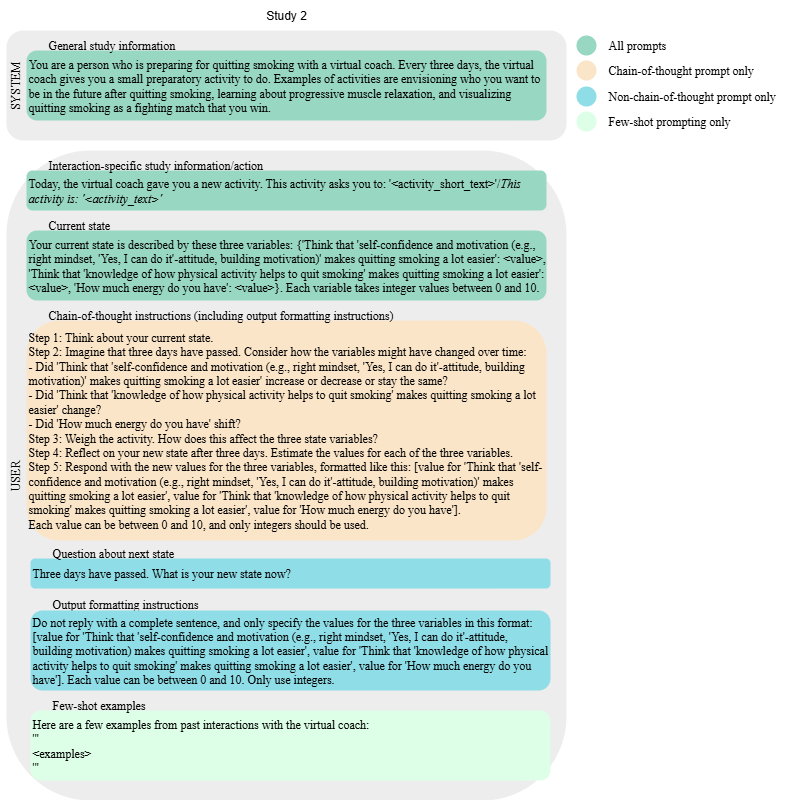}
    \caption{Prompt template for the next state for study 2 on activities for quitting smoking. The part in italics is only shown for the extensive prompt version.}
    \label{fig:prompt_template_study2_nextstate}
\end{figure}

\begin{figure}
    \centering
    \includegraphics[width=\linewidth]{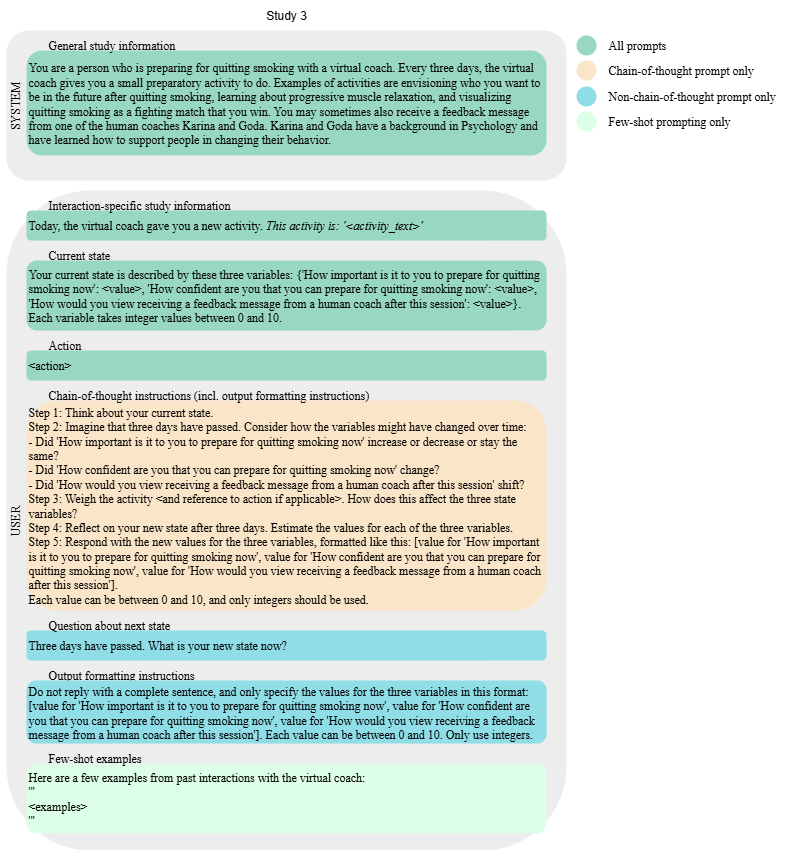}
    \caption{Prompt template for the next state for study 3 on human feedback for quitting smoking. The part in italics is only shown for the extensive prompt version.}
    \label{fig:prompt_template_study3_nextstate}
\end{figure}

\begin{figure}
    \centering
    \includegraphics[width=\linewidth]{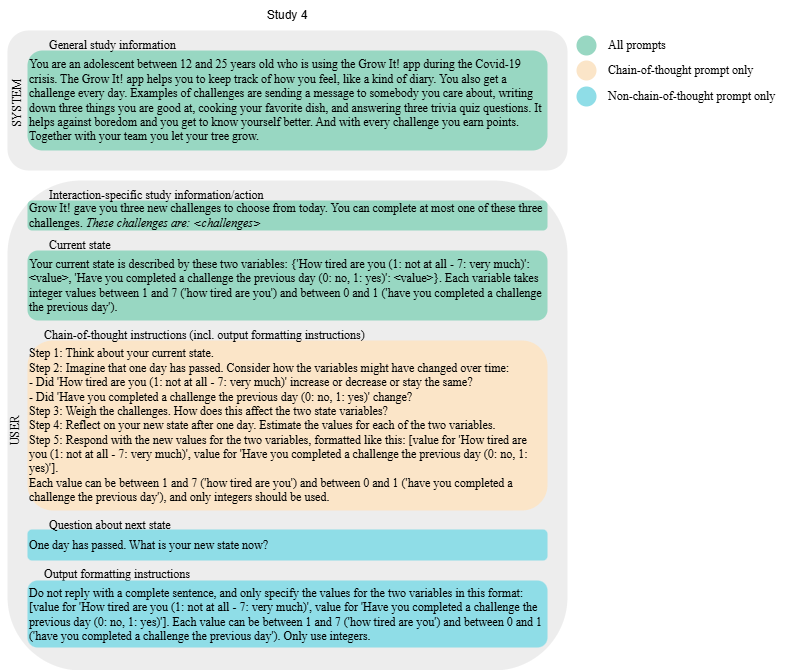}
    \caption{Prompt template for the next state for study 4 on coping challenges for mental well-being. The part in italics is only shown for the extensive prompt version, which is the only prompt version we use for this study due to the brevity of the challenge texts.}
    \label{fig:prompt_template_study4_nextstate}
\end{figure}

Fig.~\ref{fig:prompt_template_study1_reward} through Fig.~\ref{fig:prompt_template_study4_nextstate} show the prompt templates used for the reward and next state for all four studies. Note that all prompts follow a similar high-level structure, but that small changes had to be made between studies (e.g., because some studies had two and others had three state variables influencing the learned parts of the reward and transition functions).

\clearpage

\subsection*{Human-generated samples: Asking human raters to generate samples}

\begin{figure}[h!]
    \centering
    \includegraphics[width=0.7\linewidth]{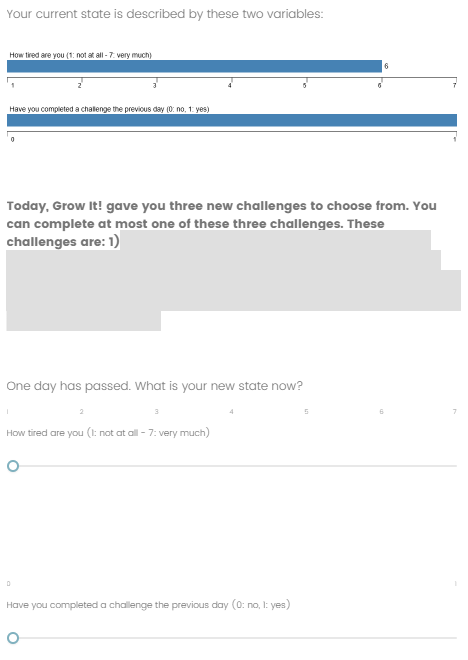}
    \caption{Screenshot of the survey in which human raters were asked to generate samples. In this specific screenshot, participants are asked to predict the next state feature values for study 4 on coping challenges for mental well-being. Beforehand, participants were given general information on the task as well as on the behavior change study they were generating samples for. The texts of the three challenges are grayed out as they cannot be publicly shared for copyright reasons.}
    \label{fig:human_survey_screenshot}
\end{figure}

\clearpage

\subsection*{Does chain-of-thought prompting help?}

\begin{table}[h]
  \caption{Example chain-of-thought reasoning by Llama-3.3-70B for the extensive prompt for study 3 when asked to predict the effort spent on the activity.}
  \label{tab:chain-of-thought_reasonings_example_reward}
  \begin{tabular}{p{12.5cm}}
 \hline
To determine the effort I'll put into the activity, I'll go through the steps as outlined:\\
\\
\#\#\# Step 1: Consider each of the variables and how they influence my effort.\\
- **Importance of preparing for quitting smoking now (5):** This moderate level of importance suggests that I recognize the need to prepare for quitting but may not be fully convinced of its immediate necessity. This could lead to a moderate level of effort, as I see some value in preparing but perhaps not enough to dedicate maximum effort.\\
- **Confidence in preparing for quitting smoking now (6):** With a slightly above moderate confidence level, I believe I have a good chance of successfully preparing for quitting, which should positively influence my effort level. I'm somewhat assured in my ability, which can motivate me to put in a significant amount of effort.\\
- **View on receiving a feedback message from a human coach (4):** This indicates a somewhat neutral or mildly positive disposition towards receiving feedback. While it might not greatly boost my motivation, it suggests I'm open to guidance and might find some value in external input, potentially enhancing my effort slightly.\\
\\
\#\#\# Step 2: Think about how the activity and the feedback message might affect me and the effort I spend on the activity.\\
- The activity itself, focusing on envisioning a future self who is more physically active, has the potential to highly motivate me. By imagining the benefits of being more active, I could become more invested in the process of quitting smoking, as it ties directly into an improved quality of life. This could significantly increase my effort.\\
- The feedback message from a human coach (either Karina or Goda) could provide an additional motivational boost, especially if it's personalized and acknowledges my current state and efforts. Their expertise in psychology could help me understand my motivations and challenges better, potentially increasing my effort by providing a sense of support and guidance.\\
\\
\#\#\# Step 3: Based on all these factors, determine how much effort you’ll put into the activity.\\
Considering the moderate importance I place on preparing for quitting, my confidence in doing so, and my openness to feedback, along with the motivational potential of the activity and the feedback message, I'm inclined to put a significant amount of effort into this activity. The activity's focus on future self-visualization and the support from a human coach could synergize with my existing motivation and confidence levels, pushing my effort towards the higher end of the scale.\\
\\
\#\#\# Step 4: Respond with a single number, representing your effort.\\
Given the considerations above, my effort would be around 8. This reflects a strong commitment to the activity, influenced by the potential motivational benefits of envisioning my future self, my current confidence and importance levels, and the supportive role of the human coach's feedback. \\
\\
effort: 8\\
  \hline
  \end{tabular}
\end{table}

\begin{table}[h]
  \caption{Example chain-of-thought reasoning by Llama-3.3-70B for the extensive prompt for study 3 when asked to predict the next state feature values.}
  \label{tab:chain-of-thought_reasonings_example_transition}
  \begin{tabular}{p{12.5cm}}
 \hline
\#\# Step 1: Understand the context of the task\\
The context is about preparing to quit smoking with the help of a virtual coach. The virtual coach provides a small activity every three days to aid in this preparation. This suggests a structured approach to quitting smoking, focusing on gradual, manageable steps towards the goal.\\
\\
\#\# Step 2: Define the current state\\
The current state is defined by three variables:\\
1. 'How important is it to you to prepare for quitting smoking now': This is currently at 0, indicating a lack of urgency or importance in preparing to quit smoking at this moment.\\
2. 'How confident are you that you can prepare for quitting smoking now': This is at 3, showing a low level of confidence in the ability to prepare for quitting.\\
3. 'How would you view receiving a feedback message from a human coach after this session': This is at 6, indicating a moderate level of openness or positivity towards receiving feedback from a human coach.\\
\\
\#\# Step 3: The activity and the feedback message\\
The activity assigned is to think about the person you want to be once you've become more physically active and to find or take a picture representing this future self. This activity aims to motivate physical activity as a step towards quitting smoking. Additionally, a feedback message from one of the human coaches, Karina or Goda, has been received, which could provide personalized encouragement or strategies based on the user's progress and needs.\\
\\
\#\# Step 4: Consider the impact of the activity and the feedback message over three days\\
Over three days, the activity is likely to increase motivation for physical activity and, by extension, for quitting smoking, as it helps in visualizing a positive future self. The feedback message could provide additional motivation, guidance, or a sense of being supported, which could boost confidence and importance attributed to quitting smoking.\\
\\
\#\# Step 5: Determine the new state\\
Given the activity and feedback, the importance of preparing to quit smoking might increase as the individual starts to see the benefits of being more physically active and the potential for a healthier lifestyle. Confidence in preparing to quit could also rise as the individual engages in activities that make them feel more capable of change. The view on receiving feedback might remain positive or increase, as the individual experiences the support and guidance it offers.\\
\\
\#\# Step 6: Format the response\\
Considering the potential impact, the new state could be:\\
- 'How important is it to you to prepare for quitting smoking now': 4 (an increase from 0, reflecting growing importance)\\
- 'How confident are you that you can prepare for quitting smoking now': 5 (an increase from 3, showing growing confidence)\\
- 'How would you view receiving a feedback message from a human coach after this session': 7 (an increase from 6, indicating a more positive view of feedback)\\
\\
The final answer is: $\boxed{[4, 5, 7]}$\\
  \hline
  \end{tabular}
\end{table}

\clearpage

\begin{figure}[h!]
 \centering
  \includegraphics[width=\textwidth]{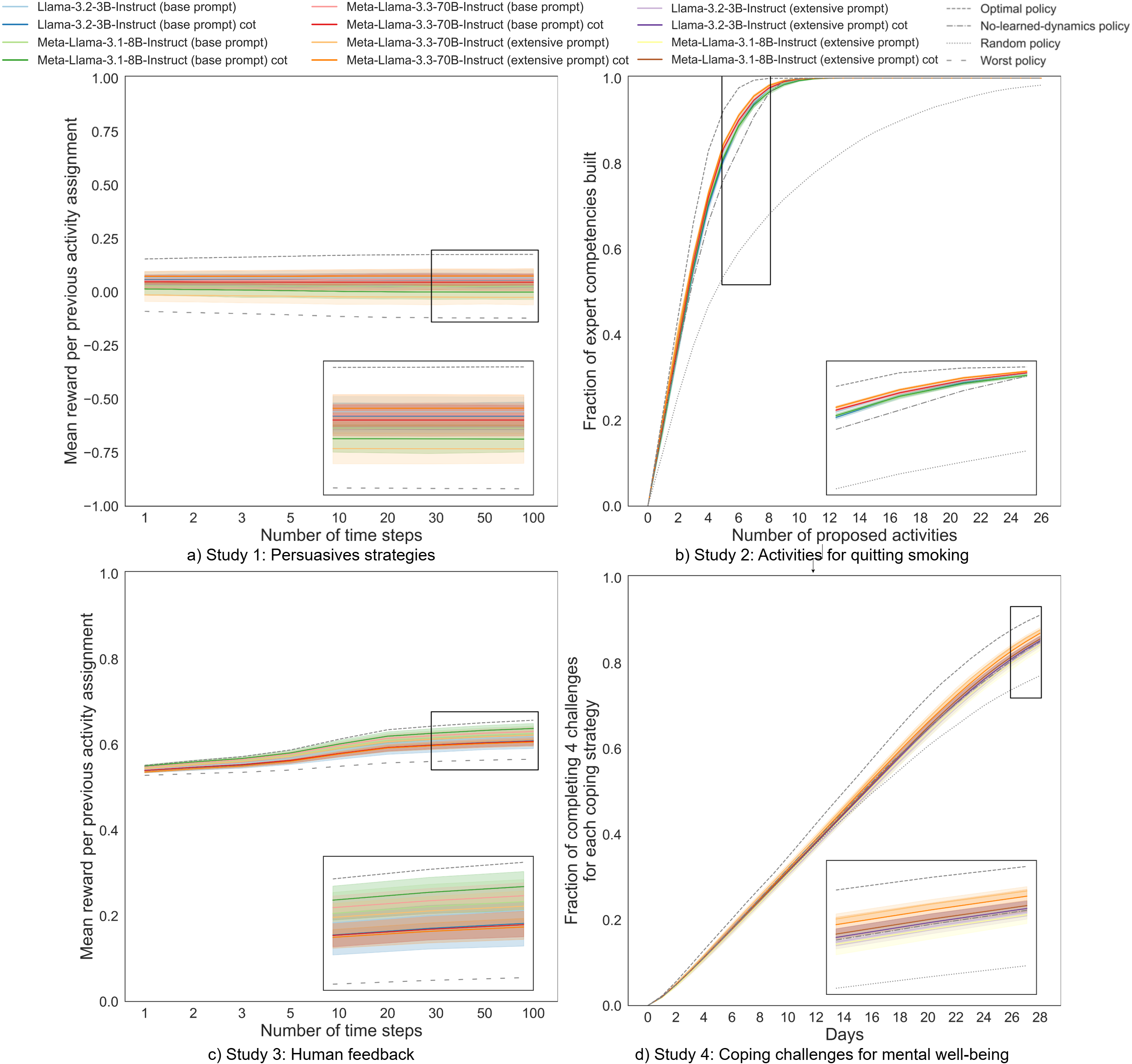}
  \caption{\small Simulated performance of policies learned from samples generated by Llama-3.3-70B-Instruct with and without chain-of-thought prompting (cot) compared to policies learned from the real behavioral samples. For each of the four behavior change studies described in Table~\ref{tab:overview_studies}, the policies are evaluated on the study-specific evaluation criterion (y-axis) over time (x-axis). The \texttt{optimal policy $\pi^*$}, \texttt{no-learned-dynamics policy}, and the \texttt{worst policy $\pi^-$} are learned based on the real behavioral data.}
  \label{fig:policy_performance_simulated_cot}
\end{figure}

\begin{figure}[h!]
 \centering
  \includegraphics[width=\textwidth]{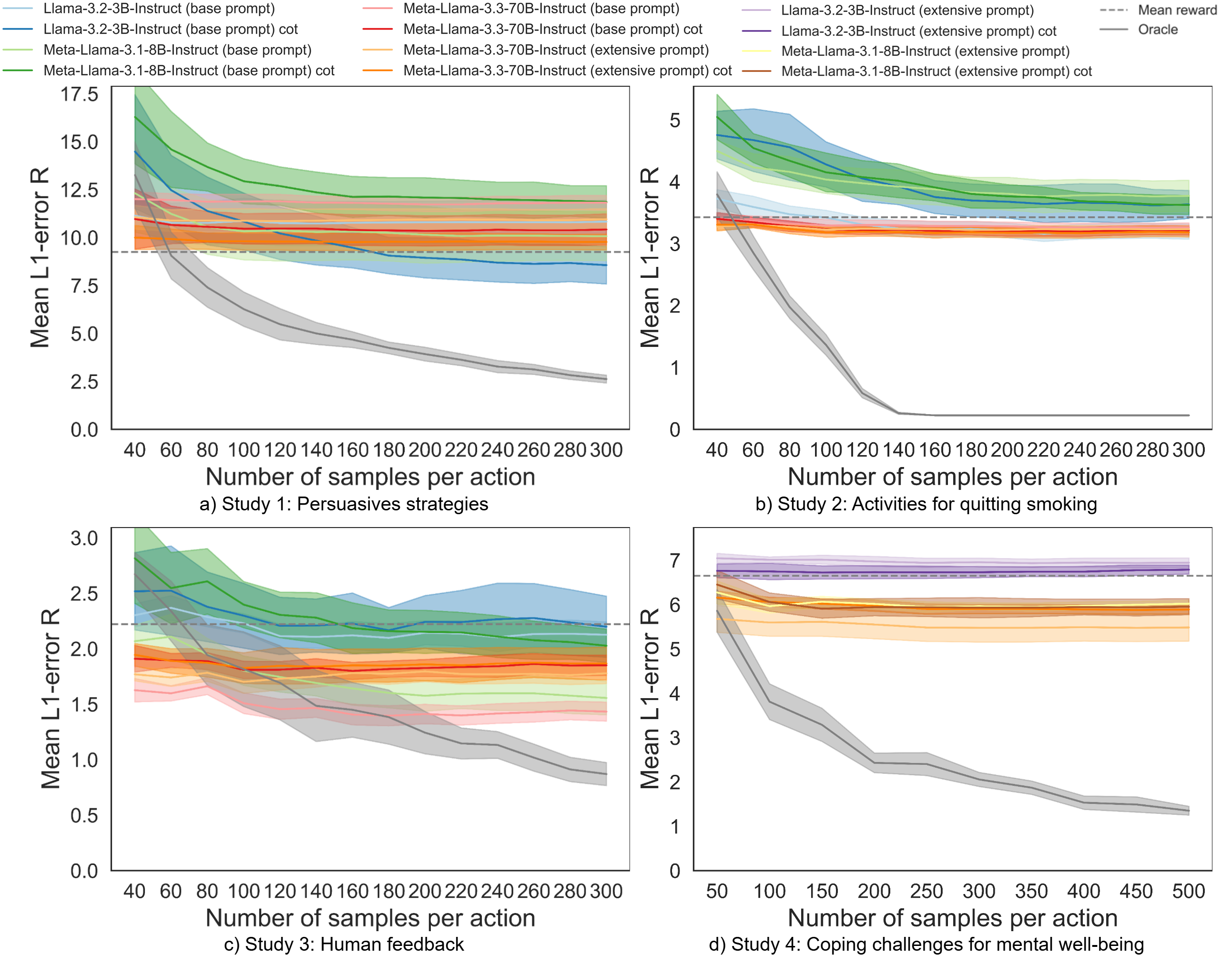}
  \caption{\small Mean $L_1$-error and 95\% credible interval between the rewards estimated from samples generated with and without chain-of-thought (cot) prompting and those estimated from the real behavioral samples for different numbers of samples per action. As comparisons serves assuming that people spend the mean reward for each state-action combination (\texttt{Mean reward}) as well as drawing samples from the real behavioral samples (\texttt{Oracle}). Means are shown over the 10 different prompt variants for LLM-generated samples and over 10 random draws for the real behavioral samples.}
  \label{fig:l1_error_rewards_cot}
\end{figure}

\begin{figure}[h!]
 \centering
 \includegraphics[width=\textwidth]{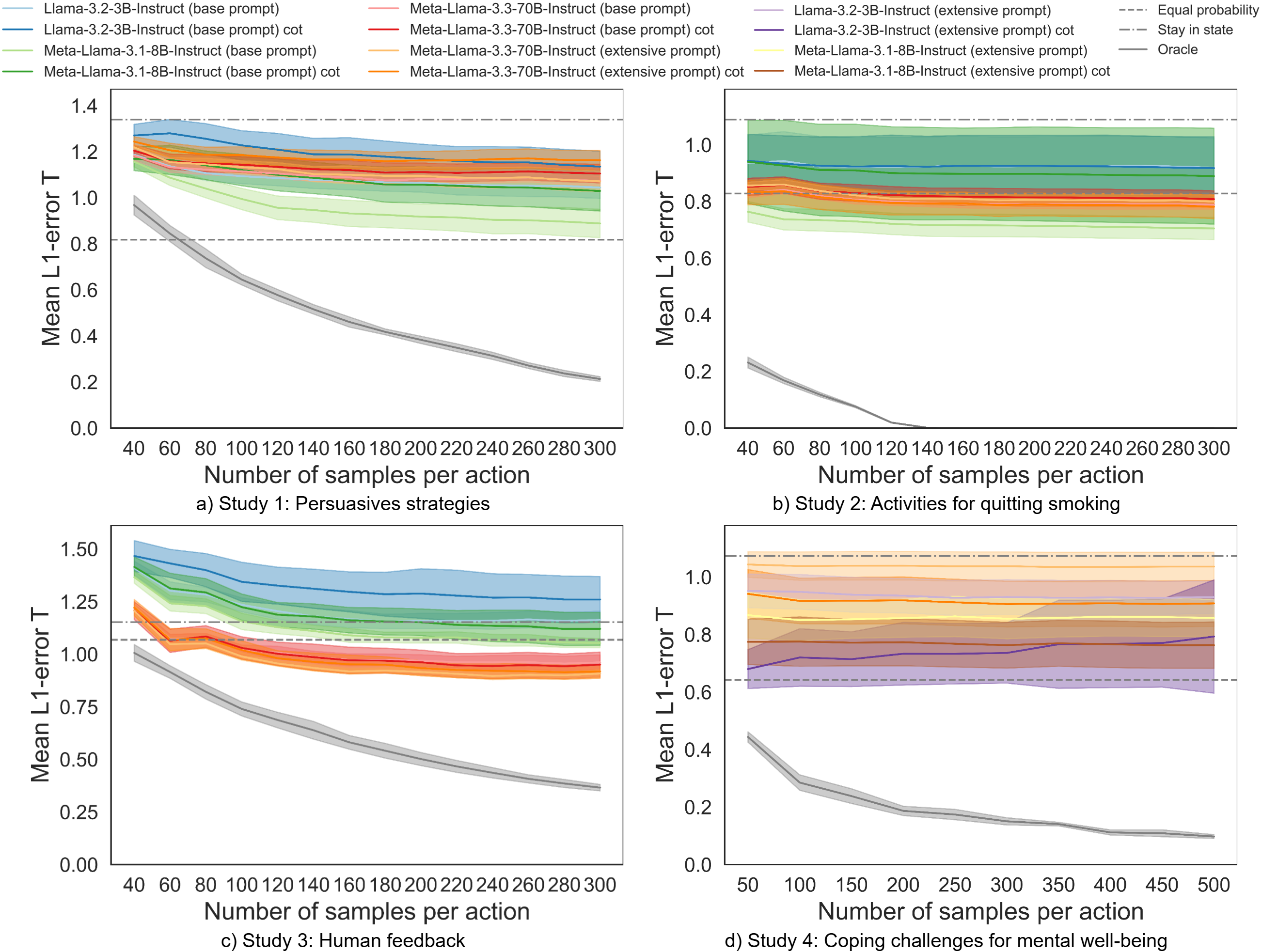}
  \caption{\small Mean $L_1$-error and 95\% credible interval between the transition functions estimated from samples generated with and without chain-of-thought (cot) prompting and those estimated from the real behavioral samples for different numbers of samples per action. As comparisons serves assuming that all transitions are equally likely (\texttt{Equal probability}), assuming that people stay in their state (\texttt{Stay in state}), and drawing samples from the real behavioral samples (\texttt{Oracle}). Means are shown over the 10 different prompt variants for LLM-generated samples and over 10 random draws for the real behavioral samples.}
\label{fig:l1_error_transitions_cot}
\end{figure}

\clearpage

\subsection*{Does changing the temperature help?}

\begin{figure}[h!]
 \centering
 \includegraphics[width=\textwidth]{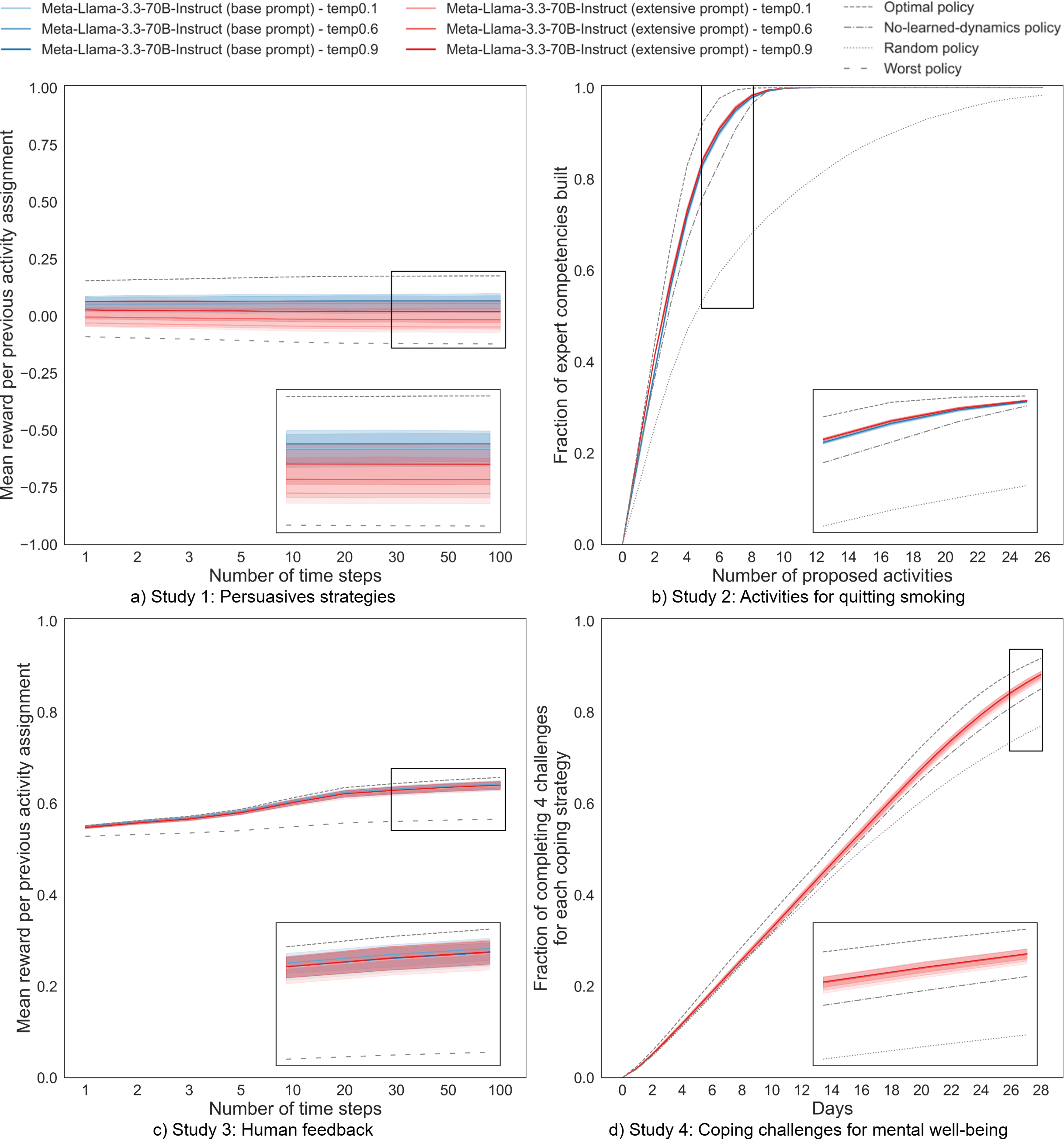}
  \caption{\small Simulated performance of policies learned from samples generated by Llama-3.3-70B-Instruct for different temperatures compared to policies learned from the real behavioral samples. For each of the four behavior change studies described in Table~\ref{tab:overview_studies}, the policies are evaluated on the study-specific evaluation criterion (y-axis) over time (x-axis). The \texttt{optimal policy $\pi^*$}, \texttt{no-learned-dynamics policy}, and the \texttt{worst policy $\pi^-$} are learned based on the real behavioral data.}
\label{fig:policy_performance_simulated_temp}
\end{figure}

\begin{figure}[h!]
 \centering
  \includegraphics[width=\textwidth]{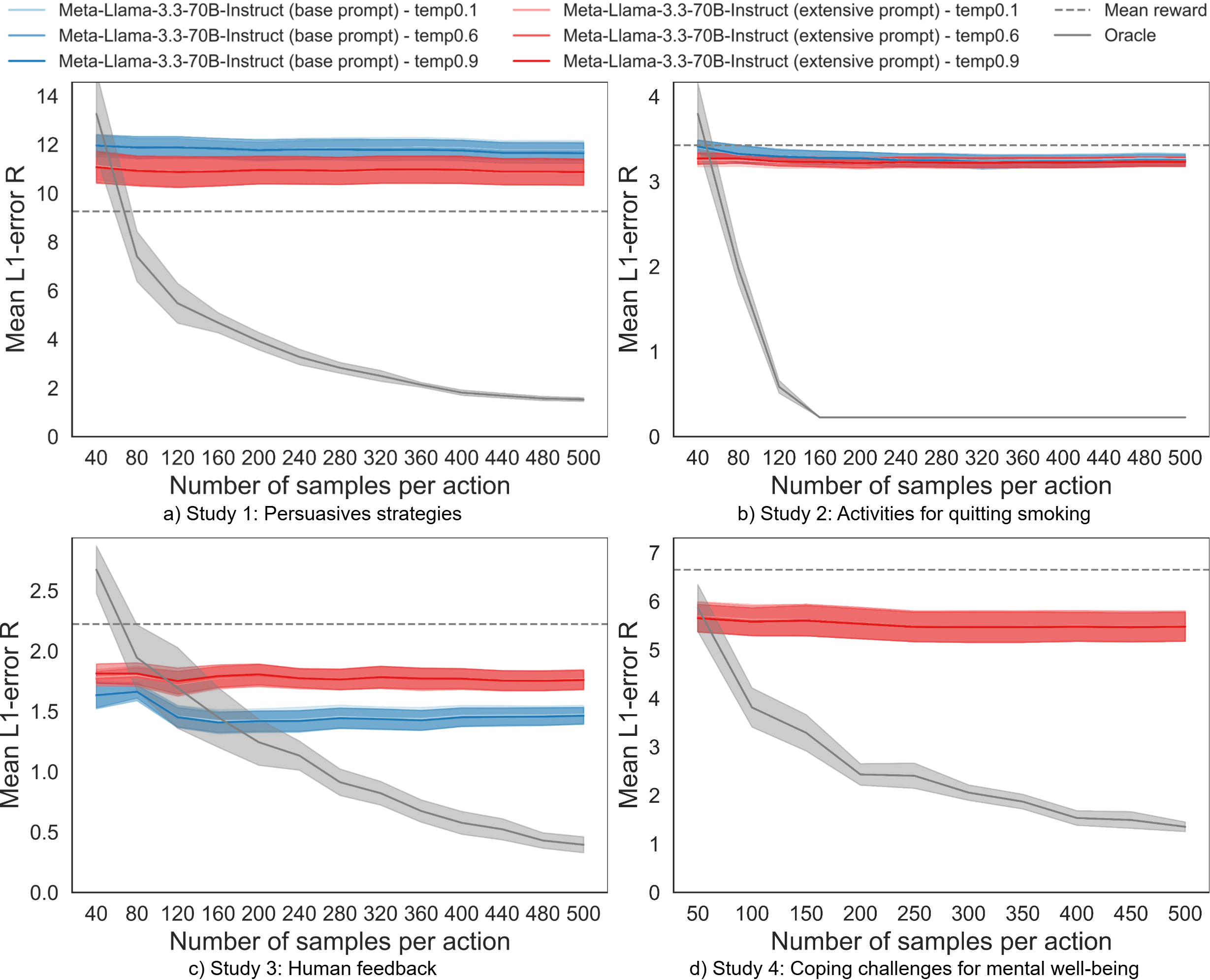}
  \caption{\small Mean $L_1$-error and 95\% credible interval between the rewards estimated from samples generated by Llama-3.3-70B-Instruct for different temperatures and those estimated from the real behavioral samples for different numbers of samples per action. As comparisons serves assuming that people spend the mean reward for each state-action combination (\texttt{Mean reward}) as well as drawing samples from the real behavioral samples (\texttt{Oracle}). Means are shown over the 10 different prompt variants for LLM-generated samples and over 10 random draws for the real behavioral samples.}
  \label{fig:l1_error_rewards_temp}
\end{figure}

\begin{figure}[h!]
 \centering
  \includegraphics[width=\textwidth]{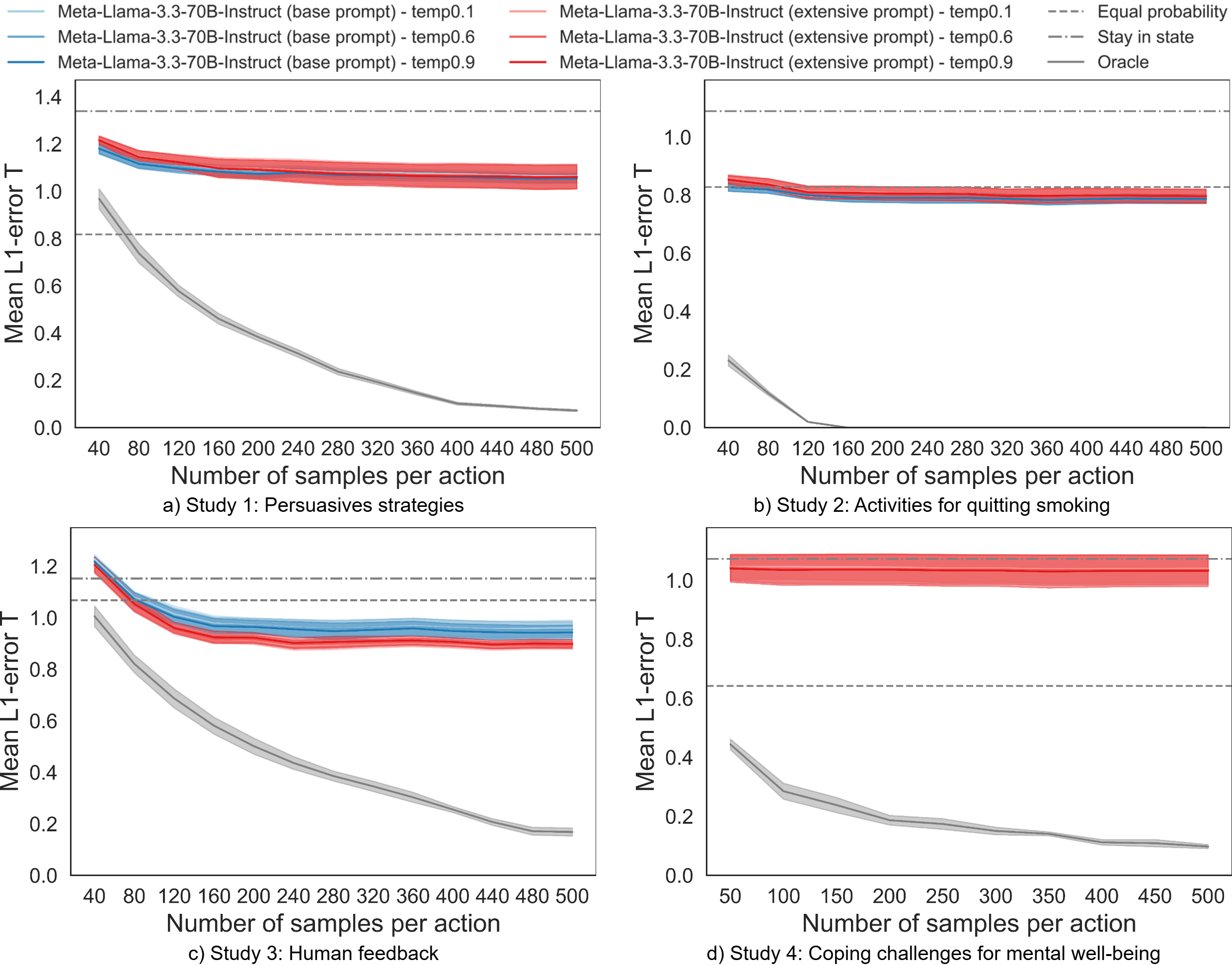}
  \caption{\small Mean $L_1$-error and 95\% credible interval between the transition functions estimated from samples generated by Llama-3.3-70B-Instruct for different temperatures and those estimated from the real behavioral samples for different numbers of samples per action. As comparisons serves assuming that all transitions are equally likely (\texttt{Equal probability}), assuming that people stay in their state (\texttt{Stay in state}), and drawing samples from the real behavioral samples (\texttt{Oracle}). Means are shown over the 10 different prompt variants for LLM-generated samples and over 10 random draws for the real behavioral samples.}
\label{fig:l1_error_transitions_temp}
\end{figure}

\clearpage

\subsection*{Does few-shot prompting help?}

\begin{figure}[h!]
 \centering
 \includegraphics[width=\textwidth]{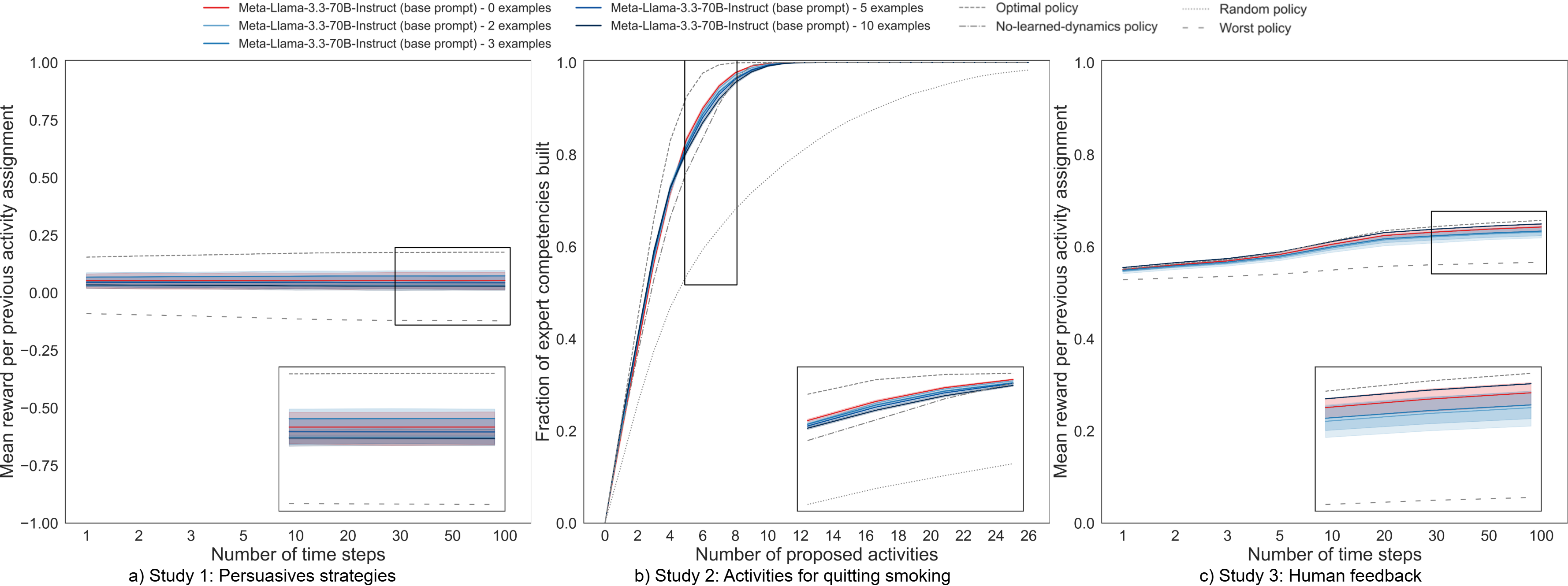}
  \caption{\small Simulated performance of policies learned from samples generated by Llama-3.3-70B-Instruct for different numbers of few-shot examples compared to policies learned from the real behavioral samples. For the first three behavior change studies described in Table~\ref{tab:overview_studies}, the policies are evaluated on the study-specific evaluation criterion (y-axis) over time (x-axis). The \texttt{optimal policy $\pi^*$}, \texttt{no-learned-dynamics policy}, and the \texttt{worst policy $\pi^-$} are learned based on the real behavioral data.}
\label{fig:policy_performance_simulated_fewshot}
\end{figure}

\begin{figure}[h!]
 \centering
  \includegraphics[width=\textwidth]{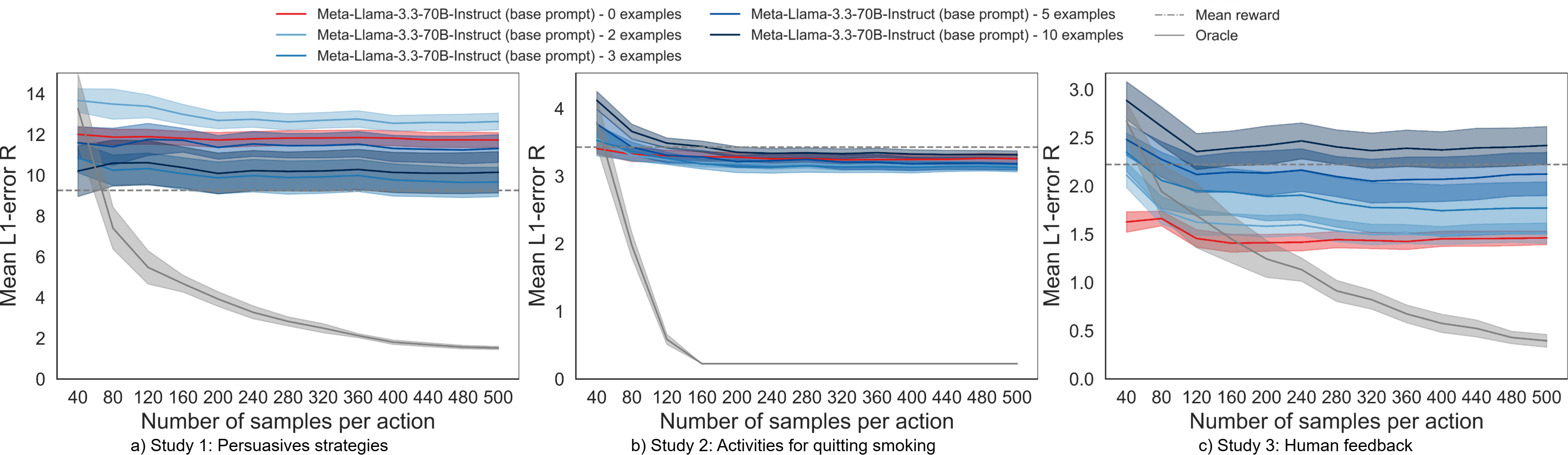}
  \caption{\small Mean $L_1$-error and 95\% credible interval between the rewards estimated from samples generated by Llama-3.3-70B-Instruct for different numbers of few-shot examples and those estimated from the real behavioral samples for different numbers of samples per action. As comparisons serves assuming that people spend the mean reward for each state-action combination (\texttt{Mean reward}) as well as drawing samples from the real behavioral samples (\texttt{Oracle}). Means are shown over the 10 different prompt variants for LLM-generated samples and over 10 random draws for the real behavioral samples.}
\label{fig:l1_error_rewards_fewshot}
\end{figure}

\begin{figure}[h!]
 \centering
  \includegraphics[width=\textwidth]{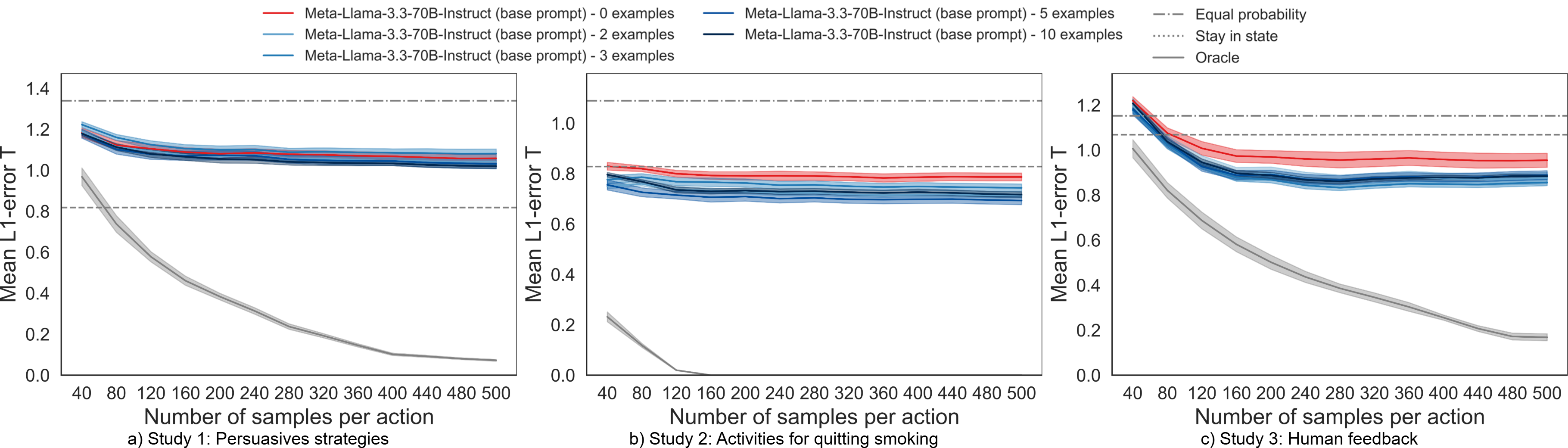}
  \caption{\small Mean $L_1$-error and 95\% credible interval between the transition functions estimated from samples generated by Llama-3.3-70B-Instruct for different numbers of few-shot examples and those estimated from the real behavioral samples for different numbers of samples per action. As comparisons serves assuming that all transitions are equally likely (\texttt{Equal probability}), assuming that people stay in their state (\texttt{Stay in state}), and drawing samples from the real behavioral samples (\texttt{Oracle}). Means are shown over the 10 different prompt variants for LLM-generated samples and over 10 random draws for the real behavioral samples.}
\label{fig:l1_error_transitions_fewshot}
\end{figure}

\end{document}